\documentclass[conference]{IEEEtran}
\IEEEoverridecommandlockouts
\usepackage{cite}
\usepackage{amsmath,amssymb,amsfonts}
\usepackage{algorithmic}
\usepackage{graphicx}
\usepackage{textcomp}
\usepackage{xcolor}

\usepackage{mathrsfs,amssymb,amsmath}
\usepackage{cite}

\usepackage{mathtools} % Bonus
\usepackage{todonotes}
\usepackage{tabularx} 
\usepackage{booktabs}
\usepackage[linesnumbered,ruled,vlined]{algorithm2e}
\SetKwInOut{Parameter}{Parameter}

\SetKwInput{KwInput}{Input}                % Set the Input
\SetKwInput{KwOutput}{Output}              % set the Output
\usepackage{hyperref}

\usepackage{tikz, pgfplots}
\usetikzlibrary{automata,positioning,arrows,arrows.meta,shapes,decorations.pathreplacing,fit,calc}
\usetikzlibrary{automata,positioning,arrows,shapes,decorations.pathreplacing,fit,calc}
\usetikzlibrary{angles,quotes}
\usetikzlibrary{decorations.pathreplacing,calc}  
\usetikzlibrary{matrix,positioning, arrows.meta}
\tikzset{signal/.style={draw,thick,minimum width=2cm,minimum height=1cm}}
\tikzset{box/.style={draw,thick,minimum width=1cm,minimum height=1cm}}
\tikzset{arrow/.style={draw,thick,->}}
\usepackage{environ}
\usepackage{placeins}
\usepackage{makecell}

\usepackage{tikz}
\usetikzlibrary{positioning}
\usetikzlibrary{decorations.pathreplacing}
\usetikzlibrary{shapes,snakes}
\usetikzlibrary{calc}
\usetikzlibrary{backgrounds}
\usepackage{adjustbox}
\usepackage{bm} 

\def\BibTeX{{\rm B\kern-.05em{\sc i\kern-.025em b}\kern-.08em
    T\kern-.1667em\lower.7ex\hbox{E}\kern-.125emX}}

\newcommand\copyrighttext{%
	\footnotesize \copyright~2024 IEEE. Personal use of this material is permitted. Permission from IEEE must be obtained for all other uses, in any current or future media, including reprinting/republishing this material for advertising or promotional purposes,creating new collective works, for resale or redistribution to servers or lists, or reuse of any copyrighted component of this work in other works.}
\newcommand\copyrightnotice{%
	\begin{tikzpicture}[remember picture,overlay]
		\node[anchor=south,yshift=10pt] at (current page.south) {\fbox{\parbox{\dimexpr\textwidth-\fboxsep-\fboxrule\relax}{\copyrighttext}}};
	\end{tikzpicture}%
}

\begin{document}

\title{Squeeze-and-Remember Block}

\author{\IEEEauthorblockN{Rinor Cakaj}
	\IEEEauthorblockA{\textit{Signal Processing} \\
		\textit{Robert Bosch GmbH \& University of Stuttgart}\\
		71229 Leonberg, Germany \\
		Rinor.Cakaj@de.bosch.com}
	\and
	\IEEEauthorblockN{Jens Mehnert}
	\IEEEauthorblockA{\textit{Signal Processing} \\
		\textit{Robert Bosch GmbH}\\
		71229 Leonberg, Germany \\
		JensEricMarkus.Mehnert@de.bosch.com}
	\and
	\IEEEauthorblockN{Bin Yang}
	\IEEEauthorblockA{\textit{ISS} \\
		\textit{University of Stuttgart}\\
		70550 Stuttgart, Germany \\
		bin.yang@iss.uni-stuttgart.de}
}

\maketitle

\begin{abstract}

Convolutional Neural Networks (CNNs) are important for many machine learning tasks. They are built with different types of layers: convolutional layers that detect features, dropout layers that help to avoid over-reliance on any single neuron, and residual layers that allow the reuse of features.
However, CNNs lack a dynamic feature retention mechanism similar to the human brain's memory, limiting their ability to use learned information in new contexts.
To bridge this gap, we introduce the ``Squeeze-and-Remember'' (SR) block, a novel architectural unit that gives CNNs dynamic memory-like functionalities. The SR block selectively memorizes important features during training, and then adaptively re-applies these features during inference. This improves the network's ability to make contextually informed predictions.
Empirical results on ImageNet and Cityscapes datasets demonstrate the SR block's efficacy: integration into ResNet50 improved top-1 validation accuracy on ImageNet by 0.52\% over dropout2d alone, and its application in DeepLab v3 increased mean Intersection over Union in Cityscapes by 0.20\%. These improvements are achieved with minimal computational overhead. This show the SR block's potential to enhance the capabilities of CNNs in image processing tasks.
\end{abstract}
\copyrightnotice

\section{Introduction}
\label{sec:intro}

\begin{figure*}[tb]
	\includegraphics[width=\linewidth]{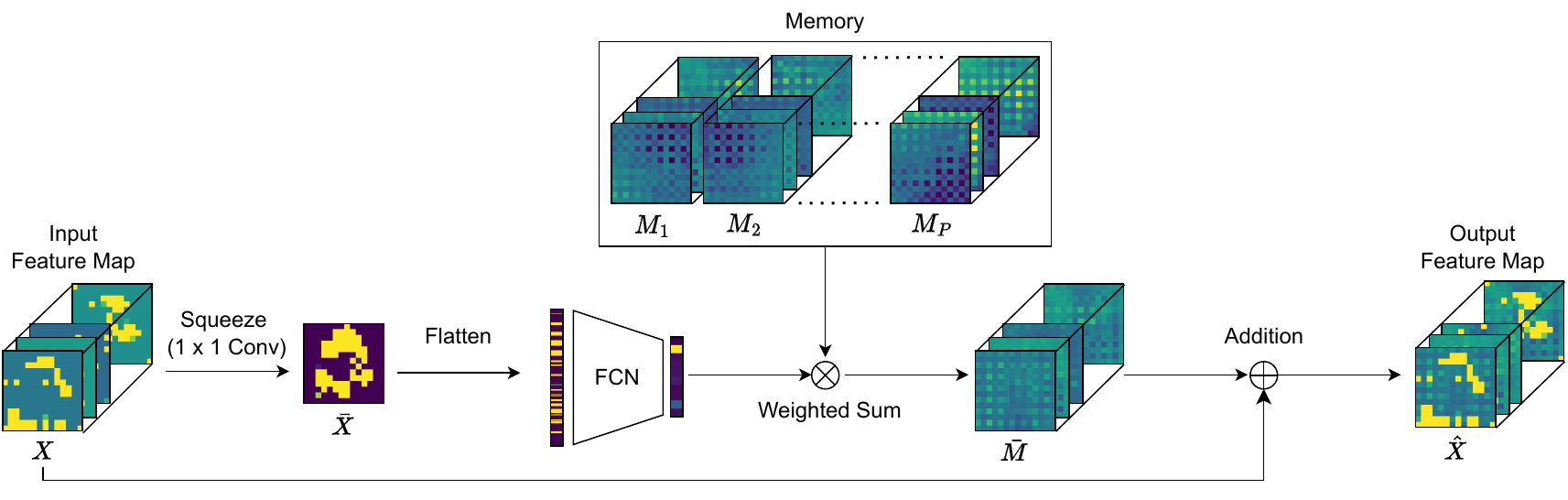}
	\caption{The Squeeze-and-Remember (SR) Block Architecture, operating in three stages: (1) ``Squeeze'': Applies a $1 \times 1$ convolution to input feature map $X$, yielding a reduced feature map $\bar{X}$ that retains essential characteristics of $X$. (2) ``Remember'': In this stage, $\bar{X}$ is flattened and passed through a Fully Connected Network (FCN) for weight calculation, followed by the computation of a weighted sum from memory blocks $M_1$ to $M_P$. These memory blocks represent a feature-spanning set, designed to capture a comprehensive range of features. The computed weighted sum, $\bar{M}$, contains high-level, possibly undetected or enhanced features such as cliff textures or architectural details. (3) ``Add'': This final step adds $\bar{M}$ to $X$, producing the output feature map $\hat{X}$.}
	\label{fig:SR_Block}
\end{figure*}

%-------------------------------------------------------------------------

Convolutional Neural Networks (CNNs) have revolutionized the field of machine learning, especially in image processing. Central to their functionality are convolutional layers, which extract a hierarchy of features from images, and fully connected layers (FC), which interpret these features for classification \cite{1989_LeCun,2018_Yamashita}. In addition, attention mechanisms, such as the Squeeze-and-Excitation (SE) block \cite{2018_Hu_CONF}, which also inspired the naming of our proposed block, refine the feature processing to further improve model performance.

Despite their advanced capabilities, CNNs still lack one crucial aspect: a memory mechanism similar to that of the human brain. Specifically, the ability to encode, store and retrieve information as seen in the hippocampus and related structures \cite{1992_Ackerman}. This gap limits their ability to dynamically use learned information.

To bridge this gap, we introduce the ``Squeeze-and-Remember'' (SR) block, a novel architectural unit for CNNs. This unit is designed to store high-level features during the training phase and then retrieve these features during inference. Such features may include the texture of a cliff or the architectural elements of a church. During inference, the SR block adaptively integrates previously undetected or relevant high-level features into feature maps, providing dynamic and context-aware addition to new inputs. 

The SR block, as shown in Figure \ref{fig:SR_Block}, begins its operation with a $1 \times 1$ convolution on the input feature map $X$. This convolution has two purposes: to reduce the depth of the feature map for computational efficiency, and to extract essential information into a compacted representation $\bar{X}$.

The compacted feature map $\bar{X}$ is then processed by a two-layer Fully Connected Network (FCN). The FCN's task is to convert this compressed feature map into a set of weights corresponding to $P$ memory blocks in the SR block's memory $M$. These memory blocks $M_i$ are designed to span a wide range of features, to ensure a comprehensive representation across the feature space. The FCN adaptively weights these blocks based on the input, ensuring that the weighted sum $\bar{M}$ contains high-level features.

The final output feature map $\hat{X}$ is obtained by adding $\bar{M}$ to the original input $X$. This fusion results in a final output that is composed of both existing and integrated features.

In our experiments, we found that the optimal placement of the SR block is within the deeper layers of CNNs. These layers, which typically process more abstract and class-relevant features, benefit from the SR block's ability to augment these features with specific, class-relevant information from its memory.

During training, the key components of the SR block: the $1 \times 1$ convolution, FCN, and memory blocks are optimized by backpropagation. This tuning process allows the block to identify and encode essential high-level features within the memory blocks. During inference, the SR block uses this training to selectively weight memory blocks based on incoming feature maps, allowing effective use of previously learned features in new contexts. The details and impact of this feature retrieval mechanism are discussed further in Section \ref{role_of_memory}.

Unlike Recurrent Neural Networks (RNNs) \cite{2020_Sherstinsky, 1997_Hochreiter}, which are highly effective in sequential data processing through feature reuse, the SR block is specifically designed for CNNs that handle static inputs such as images. It allows these networks to memorize features during training and utilize these during inference.

While weights and biases in layers such as convolutions \cite{1989_LeCun} and batch normalization \cite{2015_Ioffe_CONF} may look like they store information, they lack adaptability. Each input is processed with the same set of ``memories'', unlike the SR block, which selectively applies memorized features based on the context of the input.

Despite its advanced functionality, the SR block is computationally efficient. It easily integrates into existing CNN architectures without significant modifications and adds minimal computational overhead to the model (see Section \ref{sec:model_complexity}).

Incorporating the SR block into a ResNet50 model, using dropout2d, enhanced the top-1 validation accuracy on the ImageNet dataset by 0.52\% compared to using dropout2d alone. Additionally, for semantic segmentation tasks on the Cityscapes dataset, adding the SR block to the DeepLab v3 model, using with a ResNet50 backbone, resulted in a 0.20\% increase in mean Intersection over Union compared to the baseline model. These results highlight the effectiveness of the SR block, while maintaining a low computational footprint (see Section \ref{sec:experiments}).

\section{Related Work}
\label{sec:related_work}

This section provides an overview of the basic building blocks of Deep Neural Networks (DNNs). In particular, we will focus on CNNs, and introduce the concept of memory-augmented networks.

\subsection{Memory-Augmented Networks}

Memory-augmented networks, such as Neural Turing Machines (NTMs) \cite{2014_Graves} and Differentiable Neural Computers (DNCs) \cite{2016_Graves} are particularly effective in tasks involving sequential data, such as text and time-series analysis. These architectures give networks the ability to do complex data manipulation tasks, including copying and sorting. They achieve this by integrating neural networks with external memory components \cite{2015_Weston_CONF, 2016_Kumar_CONF}.

In contrast to these architectures, which focus on sequential tasks, our SR block offers a unique solution for non-sequential tasks, such as image classification and semantic image segmentation. The SR block diverges from traditional memory-augmented networks by learning to encode features during training, which are then adaptively recalled during inference.

% Feature Extraction
\subsection{Feature Extraction}

The convolutional layer is the cornerstone of CNNs and are used as feature extraction mechanism. In the early layers, these convolutional units detect simple patterns such as edges, corners, and curves \cite{1989_LeCun}. As the network deepens, convolutional layers progressively learn to recognize more complex and abstract features, including object classes and semantic structures \cite{2018_Yamashita}. In convolutional layers, the receptive field initially targets local regions and progressively broadens in deeper layers to cover larger image areas.

% Feature Evaluation or Classification

\subsection{Feature Evaluation or Classification}

Following feature extraction, CNNs use FC layers for feature evaluation or classification. These layers interpret the flattened feature maps from convolutional layers, classifying inputs into distinct categories \cite{2018_Yamashita}. The broad receptive field of FC layers plays an important role in evaluating extracted features to a classification output.

% Feature Weighting or Attention

\subsection{Feature Weighting or Attention}

Attention mechanisms in CNNs mirror the human cognitive process by prioritizing the most informative parts of an input. Techniques like the SE block \cite{2018_Hu_CONF} and CBAM \cite{2018_Woo_CONF} perform channel-wise and spatial feature recalibration to highlight more important features. The evolution and increasing importance of attention mechanisms \cite{2017_Vaswani_CONF} highlight their role in enhancing CNN performance.

% Feature Reuse

\subsection{Feature Reuse}

Residual and recurrent connections in DNNs allows feature reuse, giving the network the possibility to use extracted features in subsequent layers \cite{2016_He_CONF, 2020_Sherstinsky}. This is particularly important in CNNs where convolutional layers act as both feature extractors and filters. The introduction of residual connections \cite{2016_He_CONF} shows the effectiveness of this strategy in deep network architectures.

% Feature Normalization

\subsection{Feature Normalization}

% Normalization techniques
% From FixUP Paper
Feature normalization is an important aspect of DNN training \cite{2016_Ba, 2016_Ulyanov, 2018_Wu_CONF, 2023_Cakaj_CONF}. It stabilizes and accelerate the training process. Batch normalization (BN) \cite{2015_Ioffe_CONF} standardizes feature distributions within mini-batches, reducing internal covariate shifts and thus allowing smoother training dynamics.

% Feature Pooling

\subsection{Feature Pooling}

Pooling layers reduce the spatial dimensions of feature maps through operations like max pooling \cite{2010_Scherer_CONF} and average pooling \cite{1989_LeCun_CONF}. This dimensionality reduction not only contributes to translation invariance but also decreases computational load, making the network more efficient.

% Feature Hiding

\subsection{Feature Hiding}

Regularization techniques such as dropout \cite{2014_Srivastava} and dropconnect \cite{2013_Wan_CONF} prevent overfitting by randomly dropping neurons or weights during training. While this is effective in FC layers, the efficacy of dropout in convolutional layers is limited due to the spatial correlation of their activation units. This has led to the development of structured dropout methods like dropblock \cite{2018_Ghiasi_CONF} and dropout2d \cite{2015_Tompson_CONF}, which target semantic information and encourage the network to learn distributed representations.

In summary, while each of these components has individually advanced CNN capabilities, our SR block represents an innovation by introducing a memory-like mechanism for feature retention and adaptive application in non-sequential tasks like image classification. 

\section{Squeeze-and-Remember Block}
\label{sec:squeeze_remember_block}
%Introduction Text

The \emph{Squeeze-and-Remember} (SR) block introduces a novel approach to improve CNNs. It is designed to memorize high-level features during training and adaptively recall them during inference. The SR block contains three core components:

\begin{enumerate}
	\item \textbf{Squeeze:} Compresses the feature map channels to extract essential features.
	\item \textbf{Remember:} Compute the importance of memory blocks based on the compressed feature map.
	\item \textbf{Add:} Add memorized high-level features back into the original feature map.
\end{enumerate}

\subsection{Squeeze}
The initial step in the SR block compresses the input feature map $X \in \mathbb{R}^{C \times H \times W}$ to both reduce computational complexity and extract essential features. This compression is done through a $1 \times 1$ convolution. It is chosen for its efficiency in channel-wise feature compression while minimizing additional computational load. 

During training, the $1 \times 1$ convolution learns to extract key features from the feature map that best represent the input as a whole. This learning process is facilitated by initializing the weights of this convolution uniformly within the range $(-\sqrt{k}, \sqrt{k})$, where $k = 1/C$, ensuring a balanced and effective start for the training process.

The compressed feature map $\bar{X} \in \mathbb{R}^{1 \times H \times W}$, representing these extracted features, is then flattened for further processing in the ``Remember'' stage of the SR block.

\subsection{Remember}
% Theoretical Part
In this step, the flattened feature map $\bar{X}$ is processed by a two-layer FCN. The first layer linearly transforms the feature map. The subsequent layer, assigns weights to the memory blocks using a softmax activation function. The weights of the FCN are initialized using the same uniform distribution as in the ``Squeeze'' step.

In the training phase, the FCN in the ``Remember'' step plays an important role in identifying the optimal combination of memory blocks. It learns to adaptively weight and merge these blocks. This step constructs and integrates high-level relevant features into the network. 

\subsection{Add} \label{memory_blocks}

The memory component of the SR block, represented as $M$, includes $P$ memory blocks ($M_i \in \mathbb{R}^{C \times H \times W}$). These are initially zero-filled. The blocks are trained to become a feature-spanning set, which captures a wide range of class-dependent features. The weights generated by the FCN output are used to create a weighted sum of the memory blocks, resulting in the feature map $\hat{M}$. This map is then added to the original feature map $X$. This process effectively allows the network to ``remember'' learned features in a dynamical way.

The memory blocks operate as a unique feature-spanning set, with each block $M_i$ contributing to a composite feature map $\hat{M}$ based on the input-dependent weights computed by the FCN. This adaptability of the SR block to different inputs is a key innovation that moves beyond static feature extraction.

In summary, the Squeeze-and-Remember block introduces a novel way of manipulating features in CNNs, allowing networks to adaptively reuse learned features and apply them in new contexts.

\subsection{Hyperparameters} \label{hyperparameters}

The design of new CNN architectures involves important decisions about hyperparameters and layer configurations. We focused on three key aspects for the SR block: (1) the optimal integration point in the network architecture, (2) the optimal number of neurons in the hidden layer of the FCN, and (3) the optimal number of memory blocks.

ResNets \cite{2016_He_CONF} and VGGs \cite{2015_Simonyan_CONF} are structured into several modules. Our experiments suggest that the SR block is most effective when placed after the third or fourth module. More generally, this means that the SR block should be inserted in deeper layers. This positioning allows the SR block to use more abstract, class-relevant features, and add class-specific memory content to the feature maps.

For CIFAR-100, we set the FCN's hidden layer in ResNets to contain 8 neurons, while for ImageNet, we experimented with 16 and 32 neurons. The number of memory blocks was varied between 2 to 12 for CIFAR and ImageNet. We observed that the performance of the network does not necessarily correlate with a higher number of memory blocks (see Section \ref{exp_imagenet}).

\subsection{Model and Computational Complexity} \label{sec:model_complexity}

The SR block introduces additional parameters, mainly in the memory component $M$. The total number of parameters added is calculated as:
\begin{align}
	\underbrace{C}_{\text{$1\times1$-Conv}} + \underbrace{H
		\cdot W \cdot U + U \cdot P}_{\text{FCN}} +
	\underbrace{P \cdot C \cdot H \cdot W}_{\text{Memory}},
\end{align}
where $C$ represents the channel count, $H$ and $W$ the height and width of the feature map, and $U$ the number of neurons in the FCN's hidden layer. For ResNets and VGGs on CIFAR-100, the relative increase in model parameters is typically less than 1\%. For ImageNet, due to higher resolution feature maps, the increase ranges from about 1.59\% to 7.87\%. As detailed in Section \ref{sec:experiments}, the performance improvement of the SR block is due to its feature memory capability, not just the increase in parameters.

The computational overhead of the SR block is low. In comparative tests, a standard ResNet50 and a ResNet50 with the SR block on CIFAR-100 (mini-batch of 128 images) showed minimal difference in processing times (377ms vs. 379ms, respectively). For ImageNet (mini-batch of 5 images), the times were 61.74ms for the standard ResNet50 and 62.89ms for the variant with the SR block, tested on a NVIDIA Quadro RTX 3000 GPU.

This trend holds for CPU inference times as well, which are particularly relevant for embedded device applications. For a $32 \times 32$ pixel input image, the inference time increased marginally from 41.45ms for the standard ResNet-50 to 43.01ms with the SR block.

\section{Experiments}
\label{sec:experiments}

In this section, we evaluate the effectiveness of our Squeeze-and-Remember block in enhancing CNNs for supervised image recognition and semantic segmentation tasks. Our empirical studies primarily focus on two key areas: image recognition using CIFAR-100 and ImageNet datasets, and semantic segmentation on the Cityscapes dataset. We compare the performance of various CNN architectures using the SR block against their standard configurations. Additionally, we evaluate the synergistic effect of integrating the SR block with regularization techniques like dropout \cite{2014_Srivastava}, dropout2D \cite{2015_Tompson_CONF}, and dropblock \cite{2018_Ghiasi_CONF}.

For clarity and consistency, our model descriptions follow a specific naming convention. For example, ``ResNet50 + D'' denotes a ResNet50 model trained with dropout, ``D2D'' for dropout2D, and ``DB'' for dropblock. Models that include the SR block are indicated by adding ``SR'' to their name, such as ``ResNet50 + SR'' or ``ResNet50 + D2D + SR''.

\subsection{Experiments on CIFAR}

Our evaluation of the SR block was performed on the CIFAR-10/100 datasets, containing 50,000 training and 10,000 test color images of 32x32 pixels. We utilized a range of CNN architectures for our experiments, including ResNet18/34/50 \cite{2016_He_CONF} and VGG16/19-BN \cite{2015_Simonyan_CONF}. Detailed results for CIFAR-10 are provided in Appendix \ref{results_cifar_10}.

To ensure robustness and reproducibility, each model was trained and evaluated five times with different random seeds. This affects the network initialization, data ordering, and augmentation processes. We present the mean test accuracy and its standard deviation. The data split was 90\% for training and 10\% for validation. The best performing model on the validation set was selected for the final evaluation.

Baseline networks were trained using data augmentation, weight decay, early stopping, and varying dropout techniques (dropout \cite{2014_Srivastava}, dropout2d \cite{2015_Tompson_CONF}, dropblock \cite{2018_Ghiasi_CONF}). We also investigated the performance improvement across ResNet architectures when the Squeeze-and-Remember block was used alongside SE \cite{2018_Hu_CONF} and CBAM blocks \cite{2018_Woo_CONF}. The VGG models (VGG16-BN and VGG19-BN) followed a similar training regimen, with dropout applied only in their fully connected (FC) layers. Comprehensive implementation details are available in Appendix \ref{implementation_details_cifar}.

Table \ref{tab:cifar100} presents our findings on the CIFAR-100 dataset. In this table, \textbf{M} indicates the SR block's integration point, \textbf{P} the number of memory blocks, and \textbf{OH} the additional parameter overhead. Notably, all ResNet and VGG models incorporating the SR block showed consistent improvements in accuracy. For example, ResNet50 using the SR block showed an increase of $1.05\%$ in accuracy.

A key observation from our study is the improved performmance of the SR block when used together with regularization techniques like dropout, dropout2D, and dropblock. This synergistic effect suggests that the benefits of each of these methods complement each other. We think that the regularization methods may influence the SR block's memory blocks to focus on class-dependent instead of instance-specific features. This can contribute to the accuracy gains.

Importantly, the benefits of the SR block go beyond simply adding more parameters to the network. This is evident when we look at how the SR block improves performance compared to simply making the network larger, such as the more than 10\% increase in parameters when moving from ResNet34 to ResNet50. These comparisons show that the effectiveness of the SR block comes from its ability to remember and use features in a unique way, not just from adding more parameters.

\begin{table}[tb]
	\caption{Mean Test Accuracy (Test Acc.) and Standard Deviation for CIFAR-100 Experiments. ``M'' and ``P'' indicate the SR block's integration point and number of memory blocks, while ``OH'' refers to the parameter increase percentage due to the SR block. \label{tab:cifar100}}
	
	\begin{tabular*}{\columnwidth}{@{\extracolsep{\fill}}lcccc} 
		\toprule
		\textbf{Model} & \textbf{M} & $\mathbf{P}$ & \textbf{Test Acc.} & \textbf{OH} \tabularnewline
		\midrule
		R18  & && $76.47\% \pm 0.20\%$ & \tabularnewline
		R18 + D & &&  $77.09\% \pm 0.26\%$ &  \tabularnewline
		R18 + D + SR & 3 & 10  & $\mathbf{77.26\% \pm 0.11\%}$ & $1.47\%$ \tabularnewline
		R18 + D2D & && $77.59\% \pm 0.24\%$ & \tabularnewline
		R18 + D2D + SR& 4 & 12& $\mathbf{77.93\% \pm 0.25\%}$ & $0.88\%$ \tabularnewline
		R18 + DB & && $78.01\% \pm 0.20\%$ &  \tabularnewline
		R18 + DB + SR  & 3 & 2& $\mathbf{78.12\% \pm 0.15\%}$  & $0.30\%$ \tabularnewline
		R18 + SR  & 4 &2  & $76.74\% \pm 0.33\%$ & $0.15\%$ \tabularnewline
		\midrule
		R34  & && $77.07\% \pm 0.45\%$ &  \tabularnewline
		R34 + D & && $77.60\% \pm 0.49\%$ & \tabularnewline
		R34 + D + SR &4 &4& $\mathbf{77.96\% \pm 0.38\%}$ & $0.16\%$ \tabularnewline
		R34 + D2D & && $78.71\% \pm 0.28\%$ & \tabularnewline
		R34 + D2D + SR & 3&6& $\mathbf{78.81\% \pm 0.28\%}$ & $0.46\%$ \tabularnewline
		R34 + DB & && $78.68\% \pm 0.22\%$ & \tabularnewline
		R34 + DB + SR&3 &2& $\mathbf{78.84\% \pm 0.22\%}$  & $0.16\%$ \tabularnewline
		R34 + SR &3 &6& $77.22\% \pm 0.30\%$ & $0.46\%$ \tabularnewline
		\midrule
		R50  & && $76.17\% \pm 0.69\%$ & \tabularnewline
		R50 + D & && $77.47\% \pm 0.40\%$ & \tabularnewline
		R50 + D + SR&4&6& $\mathbf{77.72\% \pm 0.29\%}$ & $0.21\%$ \tabularnewline
		R50 + D2D &&& $78.58\% \pm 0.58\%$ & \tabularnewline
		R50 + D2D + SR &3&8& $\mathbf{78.68\% \pm 0.55\%}$ & $0.55\%$ \tabularnewline
		R50 + DB &&& $78.34\% \pm 0.56\%$ & \tabularnewline
		R50 + DB + SR &4&4& $\mathbf{78.43\% \pm 0.29\%}$  & $0.14\%$ \tabularnewline
		R50 + SR &3&2& $77.22\% \pm 0.74\%$ & $0.14\%$ \tabularnewline
		\midrule
		VGG16  & &&$72.48\% \pm 0.35\%$ & \tabularnewline
		VGG16 + SR &2 &8&$\mathbf{72.87\% \pm 0.27\%}$ & $0.42\%$ \tabularnewline
		VGG19  & &&$71.34\% \pm 0.14\%$ & \tabularnewline
		VGG19 + SR & 4&10&$\mathbf{71.42\% \pm 0.27\%}$ & $0.10\%$ \tabularnewline
		\bottomrule
	\end{tabular*}
\end{table}

Table \ref{tab:sr-enhancement-cifar100-combination} present the effect of adding the Squeeze-and-Remember block to SE and CBAM on CIFAR-100. It significantly improves performance across different ResNet models. For example, using the SR block with CBAM in ResNet34 boosts accuracy by 0.53\%. When the SR block is combined with the SE block in ResNet50, there's a 0.43\% increase in accuracy. 

The improvements from the SR block show its ability to act as a supplementary component that memorizes important features during training, which are then adaptively recalled during inference. This improves the ability of the network to make predictions by reintegrating high-level features, extending the scope of feature enhancement beyond recalibration.

\begin{table}[ht]
	\centering
	\caption{Mean Test Accuracy (Test Acc.) and Standard Deviation for CIFAR-100 Experiments. Columns ``M'' and ``P'' denote the integration point of the SR block and the number of memory blocks. The ``OH'' column indicates the percentage increase in parameters attributable to the use of feature-modifying blocks such as SE, CBAM, and SR.}
	\label{tab:sr-enhancement-cifar100-combination}
	\begin{tabular*}{\columnwidth}{@{\extracolsep{\fill}}lcccc} 
		\toprule
		\textbf{Model} & \textbf{M} & \textbf{P} & \textbf{Test Acc.} & \textbf{OH} \\
		\midrule
		ResNet18 + CBAM & - & - & 76.71\% $\pm$ 0.28\% & 0.78\% \\
		\textbf{ResNet18 + CBAM + SR} & 3 & 6 & 76.96\% $\pm$ 0.24\% & 1.66\% \\
		ResNet18 + SE & - & - & 76.96\% $\pm$ 0.24\% & 0.79\% \\
		ResNet18 + SE + SR & 3 & 2 & 76.86\% $\pm$ 0.21\% & 1.09\% \\
		\midrule
		ResNet34 + CBAM & - & - & 77.20\% $\pm$ 0.33\% & 0.74\% \\
		\textbf{ResNet34 + CBAM + SR} & 3 & 6 & \textbf{77.73\% $\pm$ 0.21\%} & 1.05\% \\
		ResNet34 + SE & - & - & 77.73\% $\pm$ 0.22\% & 0.76\% \\
		\textbf{ResNet34 + SE + SR} & 4 & 2 & \textbf{77.93\% $\pm$ 0.23\%} & 0.84\% \\
		\midrule
		ResNet50 + CBAM & - & - & 78.73\% $\pm$ 0.54\% & 10.61\% \\
		\textbf{ResNet50 + CBAM + SR} & 3 & 6 & \textbf{79.06\% $\pm$ 0.29\%} & 12.28\% \\
		ResNet50 + SE & - & - & 78.30\% $\pm$ 0.30\% & 10.68\% \\
		\textbf{ResNet50 + SE + SR} & 4 & 4 & \textbf{78.73\% $\pm$ 0.30\%} & 11.24\% \\
		\bottomrule
	\end{tabular*}
\end{table}

\subsection{ImageNet} \label{exp_imagenet}

We evaluated the SR block on the ImageNet 2012 classification dataset \cite{2015_Russakovsky}, which contains 1.2 million training, 50,000 validation, and 150,000 test images across 1,000 categories. Notably, regularization methods like dropout2d require longer training on ImageNet to converge effectively \cite{2018_Ghiasi_CONF}. We run our experiments four times with varied random seeds for 450 epochs using ResNet50 and report top-1 classification accuracy on the validation set. Implementation details are provided in Appendix \ref{implementation_details_imagenet}.

Table \ref{tab:imagenet} shows the top-1 validation accuracy on ImageNet. Our experiments on ResNet50 models show that incorporating the SR block with dropout2d regularization improves the network's performance, as seen by a 0.52\% increase in accuracy over the baseline with dropout2d alone. Removing the SR block from a trained ResNet50, regularized with dropout2d and using 10 memory blocks, led to a decrease of 0.4\% in accuracy. This result shows that the SR block positively impacts performance.

It is important to note that improvements are achieved even with a minimal configuration of 2 memory blocks, achieving a 0.43\% increase in accuracy. However, scaling up to 20 memory blocks does not necessarily yield proportional benefits, potentially leading to inefficiencies due to channel redundancies. Despite increasing the number of parameters, the SR block maintains a low computational overhead, preserving a small impact on inference time (see Section \ref{sec:model_complexity}).

The SR block shows its most significant performance gains when combined with dropout2d. We think that this improvement is due to the channel dropping mechanism of dropout2d, which likely encourages the memory blocks within the SR block to capture more representative, class-specific features. 

\begin{table}[tb]
	\caption{Top-1 Validation Accuracy on ImageNet. ``M'' and ``P'' indicate the SR block's integration point and number of memory blocks, while ``OH'' refers to the parameter increase percentage due to the SR block. \label{tab:imagenet}}
	
	\begin{tabular*}{\columnwidth}{@{\extracolsep{\fill}}lcccc} 
		\toprule
		\textbf{Model} & \textbf{M} & $\mathbf{P}$ & \textbf{Validation Acc.} & \textbf{OH} \tabularnewline
		\midrule
		R50  &  &&$76.87\% \pm 0.15\%$ & \tabularnewline
		R50 + D &  &&$77.53\% \pm 0.10\%$ & \tabularnewline
		R50 + D + SR &3 &2& $\mathbf{77.64\% \pm 0.10\%}$ & $1.59\%$ \tabularnewline
		\midrule
		R50 + D2D & && $77.27\% \pm 0.12\%$ & \tabularnewline
		R50 + D2D + SR &3&2& $\mathbf{77.70\% \pm 0.06\%}$ & $1.59\%$ \tabularnewline
		R50 + D2D + SR &3&10& $\mathbf{77.79\% \pm 0.02\%}$ & $7.87\%$ \tabularnewline
		R50 + D2D + SR &3&20& $77.76\% \pm 0.08\%$ & $15.73\%$ \tabularnewline
		\midrule
		R50 + DB & && $77.40\% \pm 0.11\%$ & \tabularnewline
		R50 + DB + SR &3&8& $\mathbf{77.47\% \pm 0.11\%}$  & $6.31\%$ \tabularnewline
		R50 + SR &3&10& $76.90\% \pm 0.07\%$ & $7.87\%$ \tabularnewline
		\bottomrule
	\end{tabular*}
\end{table}

\subsection{Cityscapes}

For semantic segmentation, we utilized the Cityscapes dataset \cite{2016_Cordts_CONF}, which contains high-quality, finely annotated images from 50 European cities, categorized into 19 semantic classes (2,975 training, 500 validation, and 1,525 test images). We run the experiments five times with varied random seeds using the MMSegmentation Framework \cite{mmseg2020}, with implementation notes available in Appendix \ref{implementation_details_cityscapes}. 

Table \ref{tab:cityscapes} shows the SR block's effect on semantic segmentation. In the Fully Convolutional Network (F-CONV) \cite{2015_Long_CONF} (M=2, P=3) and DeepLab v3 (DLv3) \cite{2017_Chen} (M=2, P=2), we observed improvements in Mean Intersection over Union (mIoU) and Mean Accuracy (mAcc). The SR block increased F-CONV's mIoU and mAcc by 0.2\% and 0.13\%, with 12.85\% more parameter and 1.90\% computational increase. For DLv3, mIoU and mAcc improved by 0.2\% and 0.17\%, with 24.83\% more parameters and 3.81\% more computational cost.

\begin{table}[tb]
	\caption{Comparative mIoU and mAcc Metrics on Cityscapes. This table shows the improved mean Intersection over Union (mIoU) and mean Accuracy (mAcc) achieved by incorporating the SR block into the F-CONV and DLv3 models. \label{tab:cityscapes}}
	\begin{tabular*}{\columnwidth}{@{\extracolsep{\fill}}lcc} 
		\toprule
		\textbf{Model} & \textbf{mIoU} & \textbf{mAcc} \tabularnewline
		\midrule
		F-CONV  &   $77.33\% \pm 0.16\%$ & $85.02\% \pm 0.26\%$ \tabularnewline
		F-CONV + SR & $\mathbf{77.53\% \pm 0.23\%}$ & $\mathbf{85.15\% \pm 0.29\%}$ \tabularnewline
		\midrule
		DLv3  &   $79.76\% \pm 0.32\%$ & $86.73\% \pm 0.32\%$ \tabularnewline
		DLv3 + SR & $\mathbf{79.96\% \pm 0.19\%}$ & $\mathbf{86.90\% \pm 0.23\%}$ \tabularnewline
		\bottomrule
	\end{tabular*}
\end{table}

\subsection{Interpretation of the SR mechanism} \label{role_of_memory}

\begin{figure*}
	\centering
	% Subfigure 1: Cliff
	\begin{minipage}[b]{0.30\textwidth}
		\centering
		\includegraphics[width=\textwidth]{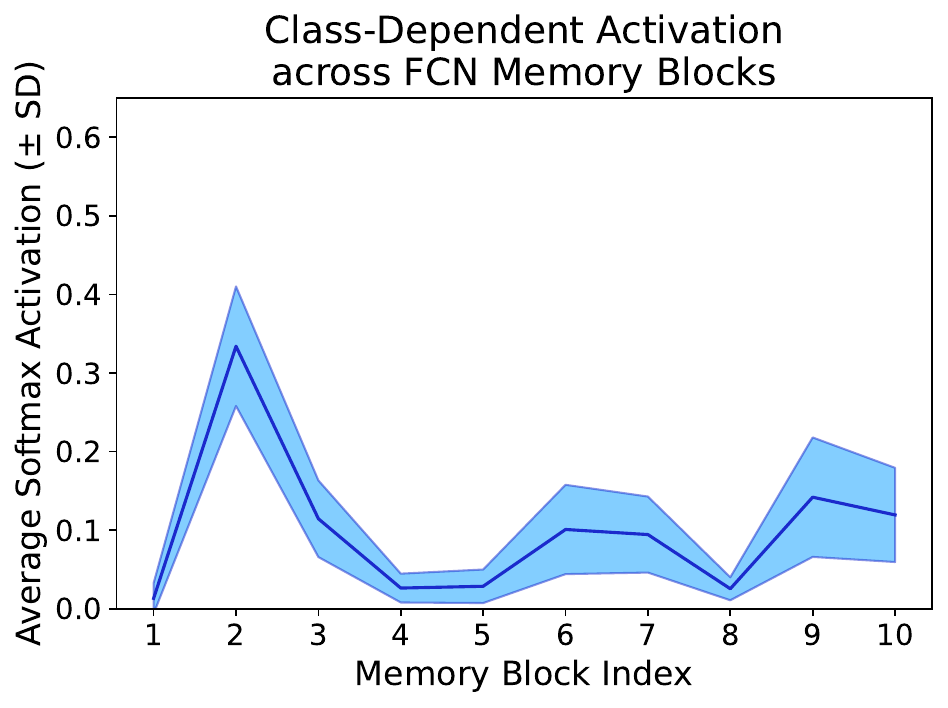}
		Class: Cliff % Manual caption without using \caption
	\end{minipage}
	\hfill
	% Subfigure 2: Pug
	\begin{minipage}[b]{0.30\textwidth}
		\centering
		\includegraphics[width=\textwidth]{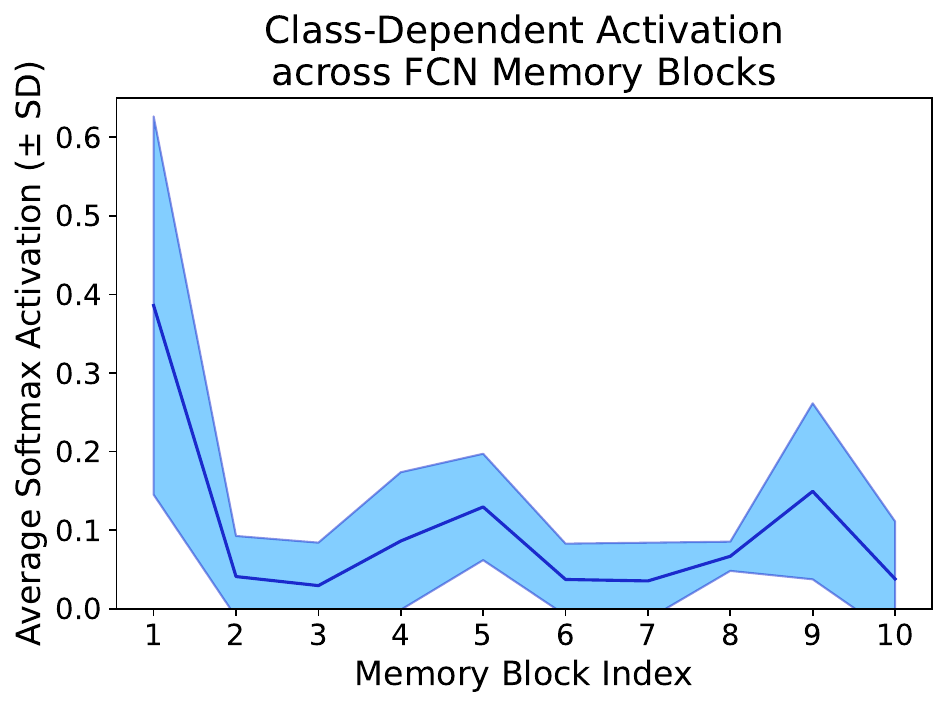}
		Class: Pug % Manual caption without using \caption
	\end{minipage}
	\hfill
	% Subfigure 3: Church
	\begin{minipage}[b]{0.30\textwidth}
		\centering
		\includegraphics[width=\textwidth]{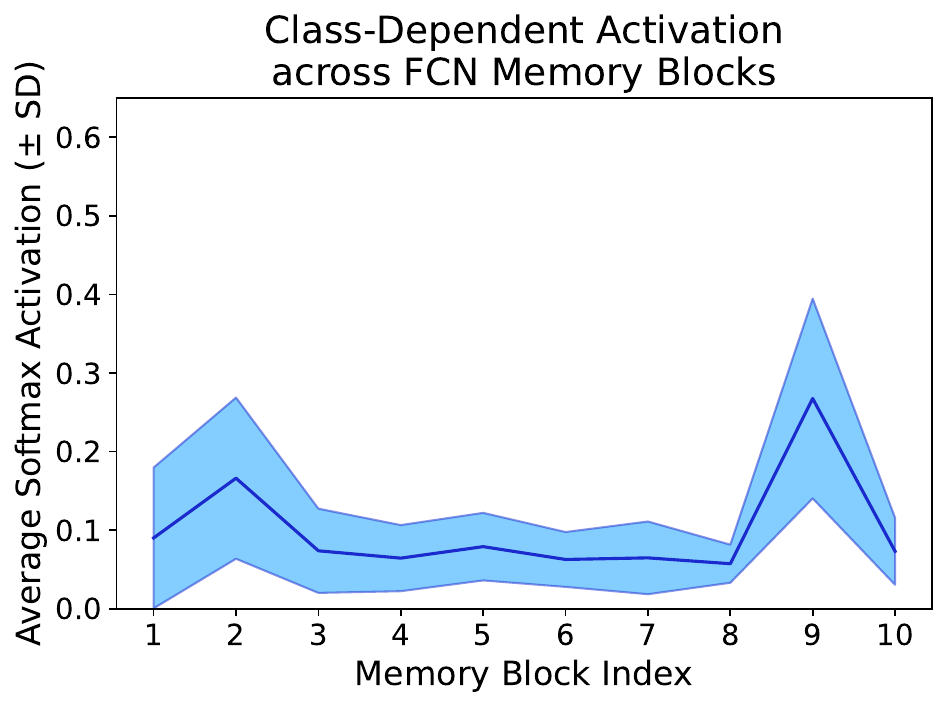}
		Class: Church % Manual caption without using \caption
	\end{minipage}
	
	\caption{Class-Conditional FCN Activation Patterns: Softmax activation means and standard deviations across ten memory blocks for ``cliff'', ``pug'', and ``church'' illustrate different strategies of memory usage. The ``cliff'' class shows diverse but consistent activation in contrast to ``pug'' and ``church''.}
	\label{fig:activations}
\end{figure*}

\begin{figure*}
	\centering
	% Subfigure 1: Cliff
	\begin{minipage}[b]{0.30\textwidth}
		\centering
		\includegraphics[width=\textwidth]{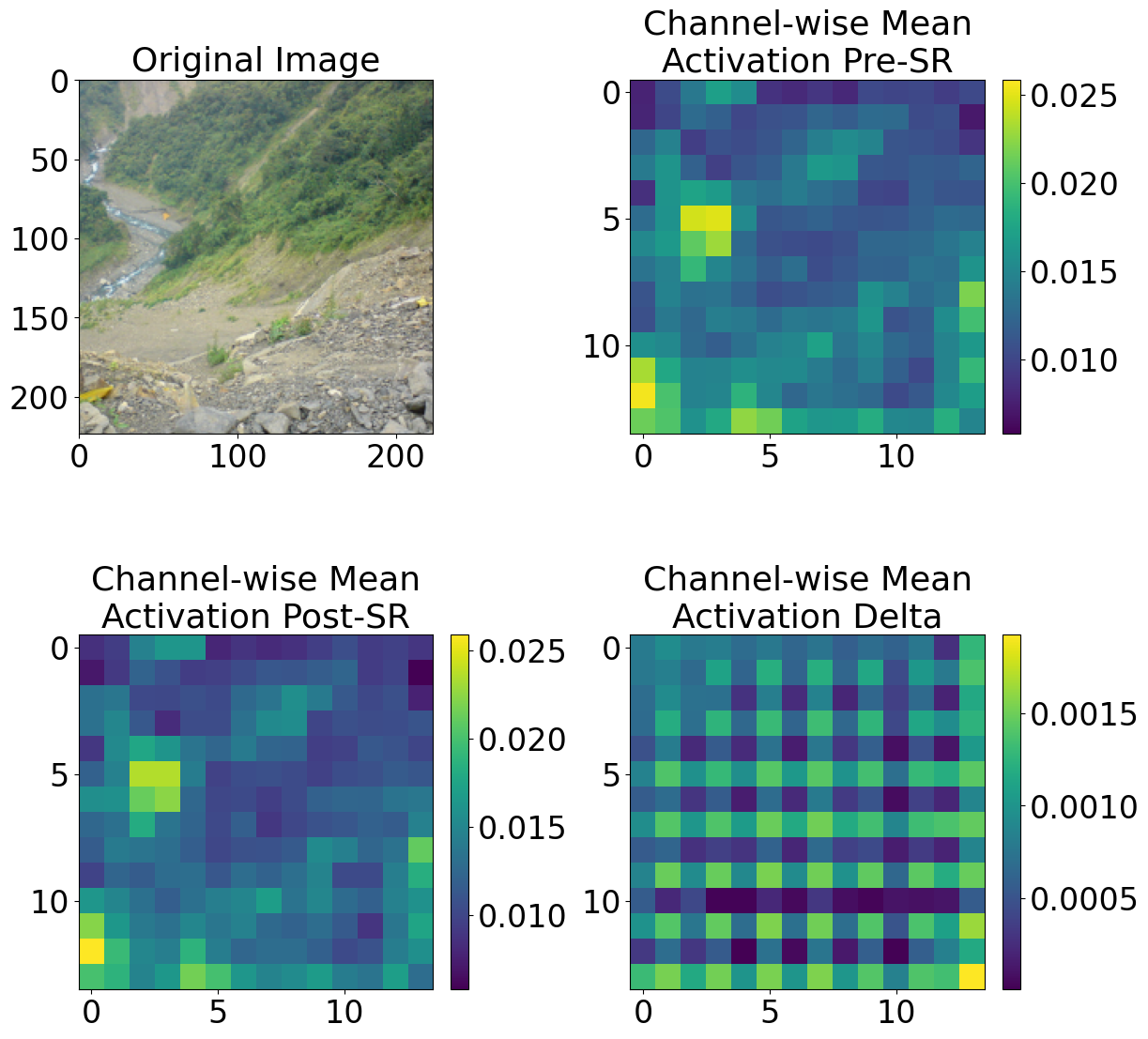}
		Class: Cliff
	\end{minipage}
	\hfill
	% Subfigure 2: Pug
	\begin{minipage}[b]{0.30\textwidth}
		\centering
		\includegraphics[width=\textwidth]{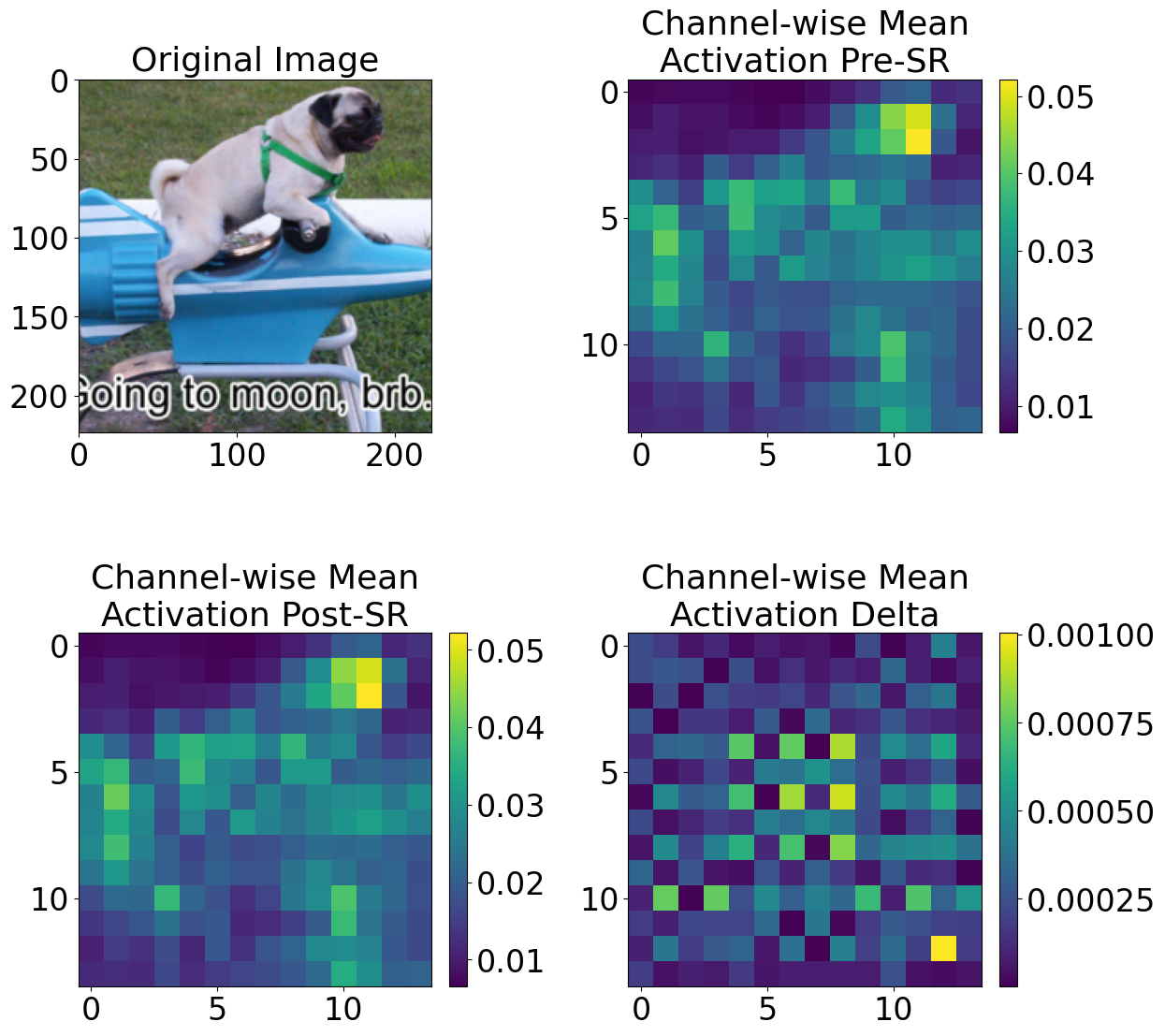}
		Class: Pug
	\end{minipage}
	\hfill
	% Subfigure 3: Church
	\begin{minipage}[b]{0.30\textwidth}
		\centering
		\includegraphics[width=\textwidth]{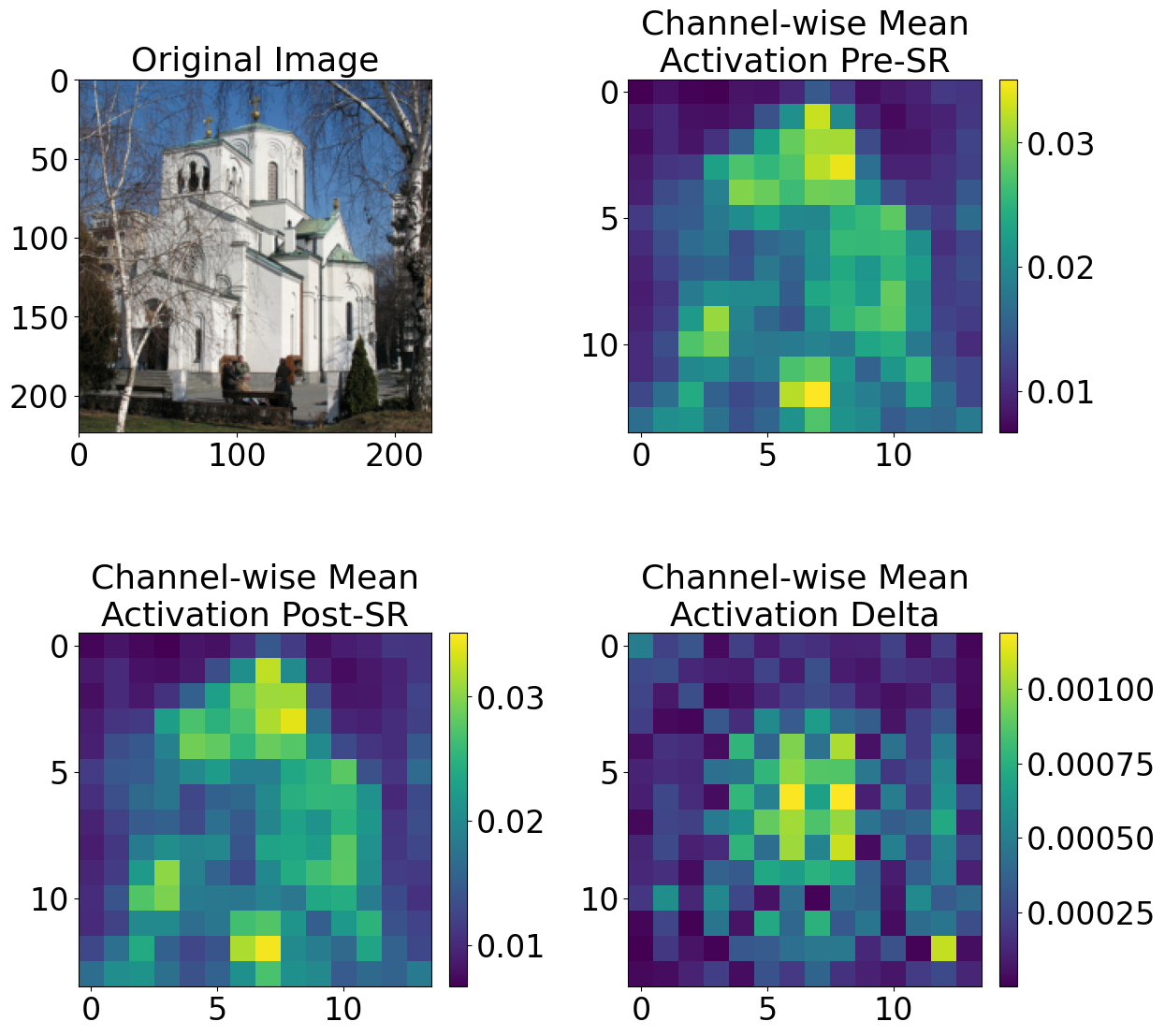}
		Class: Church
	\end{minipage}
	
	\caption{Different Impact of SR Block on Feature Maps: The figure illustrates class-dependent feature map transformations for ``Cliff'', ``Pug'', and ``Church'' classes in the SR block. It contrasts channel-wise activations before and after SR processing, with Mean Absolute Differences.}
	\label{fig:feature_maps}
\end{figure*}

\begin{figure*}
	\centering
	% Row 1
	\begin{minipage}[b]{0.18\textwidth}
		\centering
		\includegraphics[width=\textwidth]{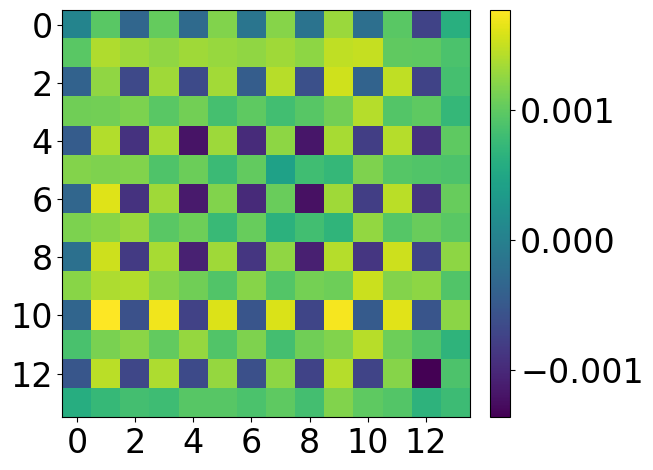}
		$M_1$
	\end{minipage}
	\hfill
	\begin{minipage}[b]{0.18\textwidth}
		\centering
		\includegraphics[width=\textwidth]{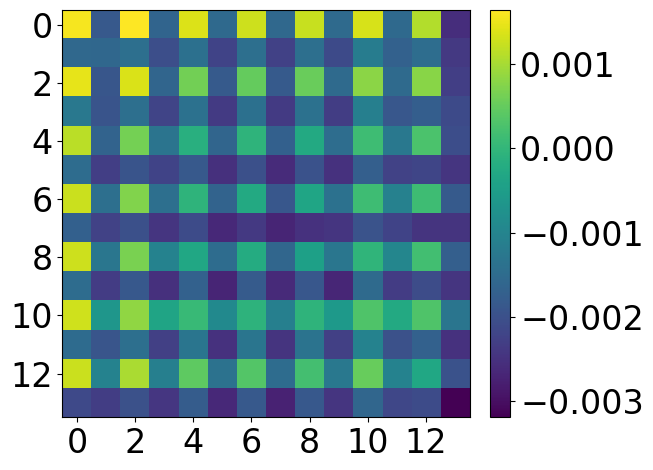}
		$M_2$
	\end{minipage}
	\hfill
	\begin{minipage}[b]{0.18\textwidth}
		\centering
		\includegraphics[width=\textwidth]{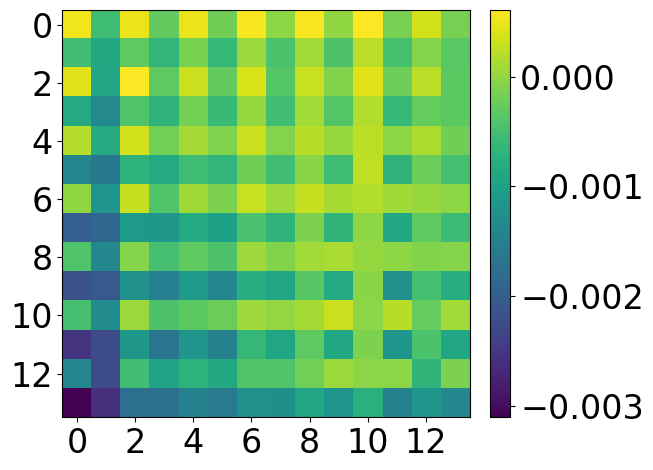}
		$M_3$
	\end{minipage}
	\hfill
	\begin{minipage}[b]{0.18\textwidth}
		\centering
		\includegraphics[width=\textwidth]{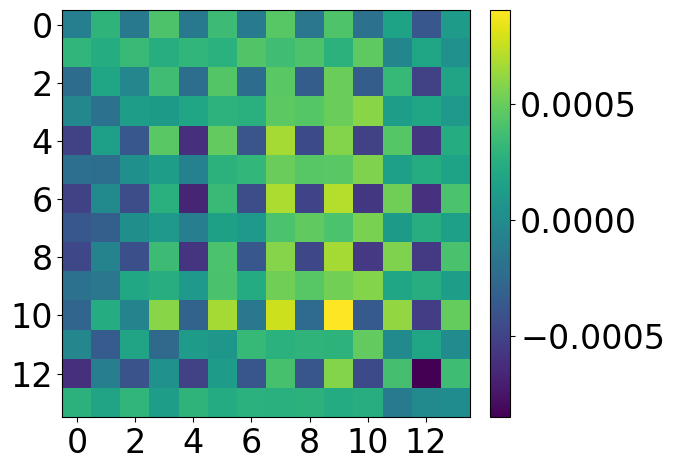}
		$M_4$
	\end{minipage}
	\hfill
	\begin{minipage}[b]{0.18\textwidth}
		\centering
		\includegraphics[width=\textwidth]{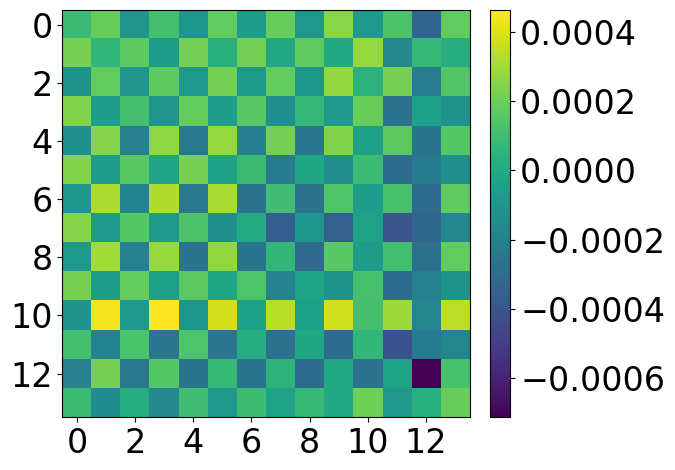}
		$M_5$
	\end{minipage}
	
	\medskip % This ensures there is a small vertical space between the two rows of images
	
	% Row 2
	\begin{minipage}[b]{0.18\textwidth}
		\centering
		\includegraphics[width=\textwidth]{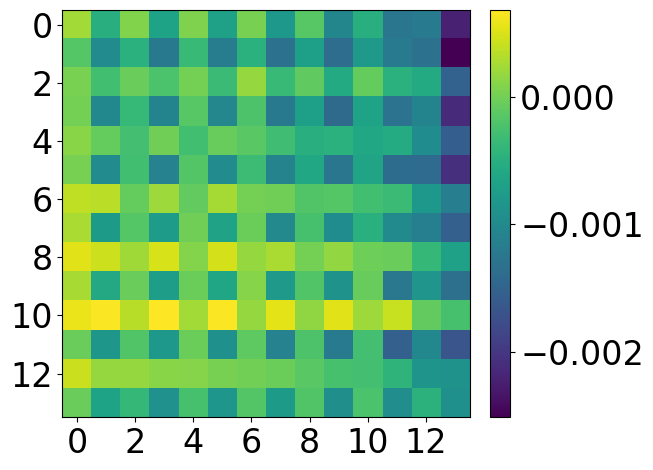}
		$M_6$
	\end{minipage}
	\hfill
	\begin{minipage}[b]{0.18\textwidth}
		\centering
		\includegraphics[width=\textwidth]{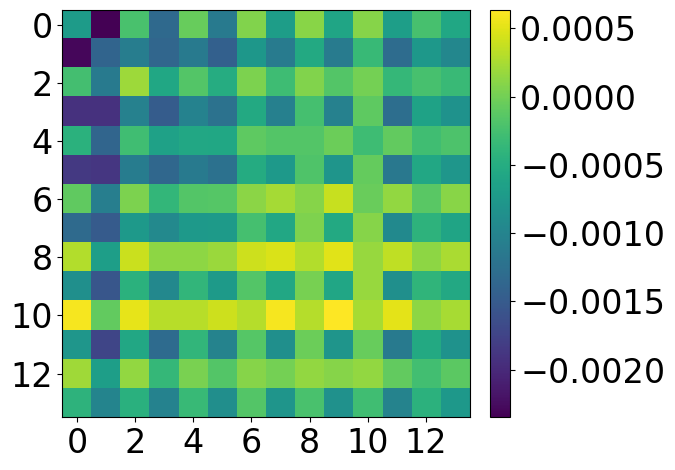}
		$M_7$
	\end{minipage}
	\hfill
	\begin{minipage}[b]{0.18\textwidth}
		\centering
		\includegraphics[width=\textwidth]{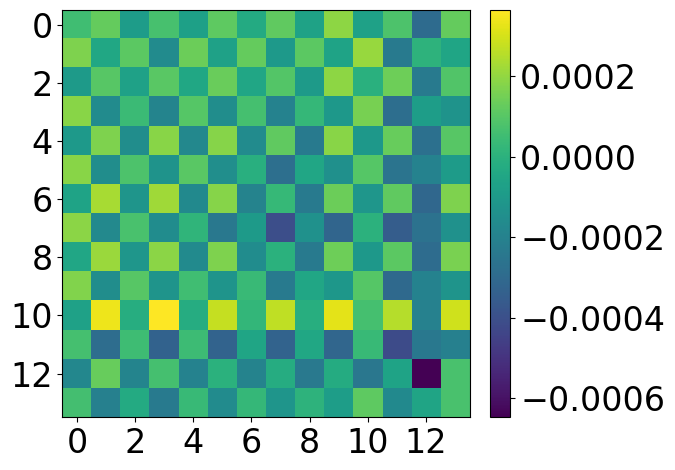}
		$M_8$
	\end{minipage}
	\hfill
	\begin{minipage}[b]{0.18\textwidth}
		\centering
		\includegraphics[width=\textwidth]{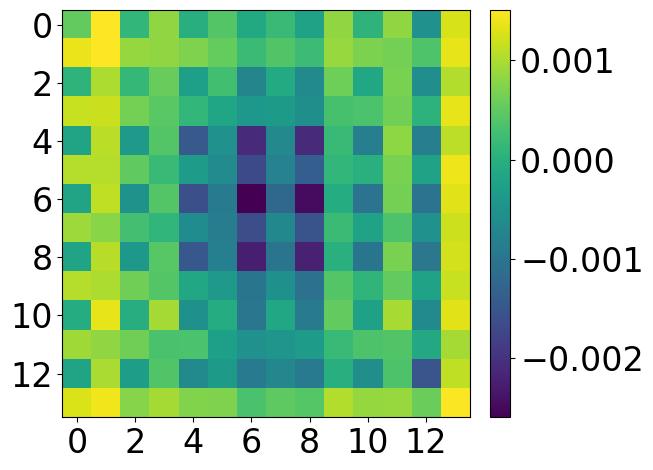}
		$M_9$
	\end{minipage}
	\hfill
	\begin{minipage}[b]{0.18\textwidth}
		\centering
		\includegraphics[width=\textwidth]{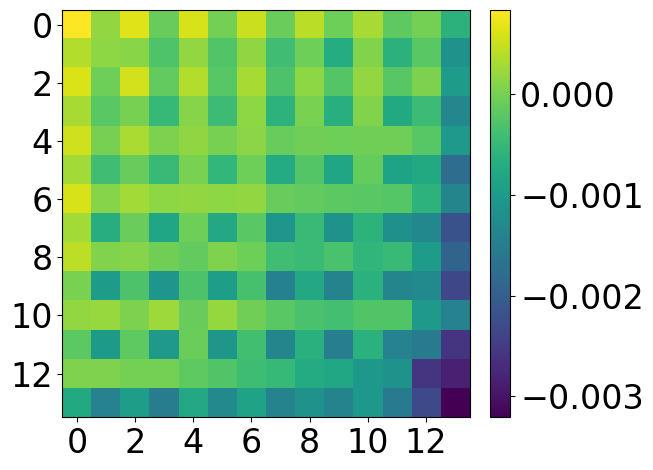}
		$M_{10}$
	\end{minipage}
	
	\caption{Feature Encoding in SR-Enhanced ResNet50: The figures $M_1$ through $M_{10}$ show the average channel activity in the ten memory blocks in the SR block. The variation in activation patterns demonstrates the memory blocks complex encoding capabilities. Each block selectively capturing different aspects of the feature spectrum.} 
	\label{fig:ch_memory_2d__0}
\end{figure*}

In this section, we investigate the functional dynamics of the SR block integrated into a ResNet50 architecture. Our analysis focuses on three areas: the softmax activation in the FCN, the influence of the SR block on the feature maps, and the characteristics of the memory blocks. The ResNet50 using a SR block with 10 memory blocks, was trained on the ImageNet dataset with dropout2d regularization.

\textbf{Softmax Activations}: We analyzed fifty samples each from the ``cliff'', ``pug'', and ``church'' classes, focusing on the mean and standard deviation of their softmax activations (see Figure \ref{fig:activations}). The results highlight different class-specific patterns in weight distribution by the FCN, showing a preference for using multiple memory blocks in the recall process. Notably, the ``cliff'' class shows a varied but consistent use of several blocks. This pattern differs from ``pug'' or ``church'' classes. This approach of combining memory blocks, allows effective high-level feature construction.

\textbf{Feature Map Modifications}: Figure \ref{fig:feature_maps} shows the SR block's ability to modify feature maps. Pre- and Post-SR channel-wise mean activations, along with their Mean Activation Delta, reveal the class-specific feature additions by the block. The Mean Activation Delta shift by 2\% to 10\%, indicating an adaptive addition of high-level features. These can be more pronounced textures in ``cliff'' landscapes or architectural features in ``church'' images.

\textbf{Memory Block Analysis}: Figure \ref{fig:ch_memory_2d__0} presents the mean channel activity for each of the SR block's ten memory blocks. By computing the channel means within each memory block, we observe a variety of activation patterns. These patterns suggest that the memory blocks represent a comprehensive feature-spanning set. The SR block combines these blocks to recall high-level feature for each class.

\subsection{Comparative Analysis with SE/CBAM Blocks}

This section compares the Squeeze-and-Remember block against the Squeeze-and-Excitation \cite{2018_Hu_CONF} and Convolutional Block Attention Module \cite{2018_Woo_CONF}. It highlights the unique contribution of the SR block to dynamic feature addition in CNNs.

\subsubsection{Initial Feature Processing}
The SE block uses global average pooling to distill spatial information into a global channel descriptor. As an extension, CBAM introduces spatial attention, which allows networks to focus on important features across both channel and spatial dimensions. In contrast, the SR block utilizes a $1 \times 1$ convolution to extract essential features for the subsequent ``Remember'' phase. This step focuses on feature extraction over aggregation.

\subsubsection{From Channel Reweighting to Adaptive Feature Recall}

The SE block recalibrates feature maps by computing weights through a bottleneck structure with sigmoid activation, based on the global context of input features. CBAM extends this by incorporating sequential channel and spatial attention mechanisms. In contrast, the ``Remember'' stage of the SR block utilizes an FCN not for feature recalibration, but for adaptively weighting and merging memory blocks to construct and reintegrate high-level features.

\subsubsection{Feature Recalibration vs Feature Integration}
SE and CBAM scale the original feature map with learned importance weights to highlight significant features. Conversely, the SR block adds memorized features through its ``Add'' operation. This goes beyond the scope of feature recalibration.

\section{Conclusion and Future Work}
\label{sec:conclusion_futuer_work}

In this work, we introduced the Squeeze-and-Remember block, an innovative architectural unit designed to improve the feature-recall capabilities of CNNs. The SR block allows a network to adaptively add memorized features into feature maps.

Incorporating the Squeeze-and-Remember block into CNNs improves performance on benchmarks like CIFAR-100 and ImageNet. A ResNet50 model with the SR block outperforms the same model with just dropout2d by 0.52\% in top-1 validation accuracy on ImageNet. Additionally, our findings highlight the SR block's effectiveness in boosting accuracy by up to 0.54\% in ResNet models when used with feature-modifying mechanisms like SE and CBAM. This shows its value as a component that improves feature processing beyond recalibration. In semantic segmentation, where the SR block is added to a ResNet50-backed DeepLab v3 model, increases the mean Intersection over Union by 0.20\% on Cityscapes. Notable, this is achieved with low extra computational demand.

Future directions include the development of training strategies that more efficiently train the memory blocks with important features. Addressing current limitations, such as the zero-initialization of memorization blocks, could further increase the impact of the SR block.

\bibliography{sr_block}
\bibliographystyle{IEEEtran}

\section{Evaluating SR Block on CIFAR-10} \label{results_cifar_10}

In complement to our main manuscript, we present extended experiments on the CIFAR-10 dataset, consisting of 50,000 training images and 10,000 test images of 32x32 pixel resolution. We utilized a range of CNN architectures for our experiments, including ResNet18/34/50 \cite{2016_He_CONF} and VGG16/19-BN \cite{2015_Simonyan_CONF}. 

To ensure robustness and reproducibility, each model was trained and evaluated five times using different random seeds, impacting network initialization, data ordering, and augmentation processes. We present the mean test accuracy and its standard deviation for these trials. The data split comprised 90\% for training and 10\% for validation, with the best-performing model on the validation set chosen for the final evaluation.

Baseline networks, including various ResNet configurations, were subjected to data augmentation, weight decay, early stopping, and varying dropout techniques (dropout, dropout2d, dropblock). The VGG models (VGG16-BN and VGG19-BN) followed a similar training regimen, with dropout applied exclusively in their fully connected (FC) layers. Detailed implementation details are available in Appendix \ref{implementation_details_cifar}.

Our CIFAR-10 experiments show consistent accuracy improvements across various models, similar to our CIFAR100 findings. The performance enhancements from adding the SR block are detailed in Table \ref{tab:cifar_10_app}, including where it's integrated (\textbf{M}), the number of memory blocks used (\textbf{P}), and the additional parameters it introduces (\textbf{OH}). Importantly, combining the SR block with regularization techniques boosts performance further than using these methods on their own.

\begin{table} [h]
	\caption{Mean Test Accuracy (Test Acc.) and Standard Deviation for CIFAR-10 Experiments. ``M'' and ``P'' indicate the SR block's integration point and number of memory blocks, respectively, while ``OH'' refers to the parameter increase percentage due to the SR block. } \label{tab:cifar_10_app}
	\begin{tabular*}{\columnwidth}{@{\extracolsep{\fill}}lcccc} 
		\toprule
		\textbf{Model} & \textbf{M} & $\mathbf{P}$ & \textbf{Test Acc.} & \textbf{OH} \tabularnewline
		\midrule
		R18  & &&$94.04\% \pm 0.08\%$ & \tabularnewline
		R18 + D & &&$94.29\% \pm 0.13\%$ & \tabularnewline
		R18 + D + SR &3&6& $\mathbf{94.36\% \pm 0.17\%}$ & $0.89\%$ \tabularnewline
		R18 + D2D &&& $94.40\% \pm 0.03\%$ & \tabularnewline
		R18 + D2D + SR &3&10& $\mathbf{94.56\% \pm 0.13\%}$ & $1.48\%$ \tabularnewline
		R18 + DB &&& $94.43\% \pm 0.14\%$ & \tabularnewline
		R18 + DB + SR &4&6&$\mathbf{94.52\% \pm 0.17\%}$  & $0.45\%$ \tabularnewline
		R18 + SR &4&4& $93.93\% \pm 0.37\%$ & $0.34\%$ \tabularnewline
		\midrule
		R34  & &&$93.69\% \pm 0.30\%$ & \tabularnewline
		R34 + D & &&$94.55\% \pm 0.31\%$ & \tabularnewline
		R34 + D + SR &3&10&$\mathbf{94.62\% \pm 0.29\%}$ & $0.77\%$ \tabularnewline
		R34 + D2D & &&$94.30\% \pm 0.14\%$ & \tabularnewline
		R34 + D2D + SR &3&8& $\mathbf{94.50\% \pm 0.12\%}$ & $0.23\%$ \tabularnewline
		R34 + DB &&& $94.38\% \pm 0.29\%$ & \tabularnewline
		R34 + DB + SR &4&2&$\mathbf{94.60\% \pm 0.21\%}$  & $0.08\%$ \tabularnewline
		R34 + SR &3&8& $94.18\% \pm 0.73\%$ & $0.62\%$ \tabularnewline
		\midrule
		R50  & &&$93.31\% \pm 0.36\%$ & \tabularnewline
		R50 + D &&& $94.09\% \pm 0.16\%$ & \tabularnewline
		R50 + D + SR &3&2& $\mathbf{94.16\% \pm 0.23\%}$ & $0.15\%$ \tabularnewline
		R50 + D2D &&& $93.99\% \pm 0.29\%$ & \tabularnewline
		R50 + D2D + SR &3&6& $\mathbf{94.01\% \pm 0.29\%}$ & $0.42\%$ \tabularnewline
		R50 + DB &&& $93.97\% \pm 0.31\%$ & \tabularnewline
		R50 + DB + SR &3&2& $\mathbf{94.21\% \pm 0.19\%}$  & $0.15\%$ \tabularnewline
		R50 + SR &3&6& $93.79\% \pm 0.41\%$ & $0.42\%$ \tabularnewline
		\midrule
		VGG16  & &&$93.27\% \pm 0.12\%$ & \tabularnewline
		VGG16 + SR &4&8&$\mathbf{93.53\% \pm 0.24\%}$ & $0.10\%$ \tabularnewline
		VGG19  & &&$93.21\% \pm 0.08\%$ & \tabularnewline
		VGG19 + SR &2&12&$\mathbf{93.39\% \pm 0.15\%}$ & $0.48\%$ \tabularnewline
		\bottomrule
	\end{tabular*}
\end{table}

\section{Detailed Methodological Framework}

\subsection{Methodology for CIFAR Datasets} \label{implementation_details_cifar}

Our CIFAR experiments used PyTorch 1.10.1 \cite{2019_Paszke_CONF} on an Nvidia GeForce 1080Ti GPU. We used the same training setup for all models, with a batch size of 128 and the SGD optimizer with a momentum of 0.9. The initial learning rate was set to 0.1, with adjustments based on the dataset: for CIFAR-10, it was reduced by a factor of 0.1 at epochs 90 and 136, along with a weight decay of $1e-4$ \cite{2016_He_CONF}; for CIFAR-100, it was reduced by 0.2 at epochs 60, 120, and 160, using a weight decay of $5e-4$. These parameter settings match those recommended in \cite{2016_He_CONF}, without further hyperparameter optimization. Data augmentation techniques included random cropping (size 32 with padding 4), random horizontal flipping, and normalization, as outlined in \cite{2019_Shorten}, with normalization being the only preprocessing step for the test set. Network initialization followed the Kaiming uniform approach \cite{2015_He_CONF}, with VGG's linear layers similarly initialized and its convolutional layers set to a Gaussian distribution ($\mu=0$, $\sigma=\sqrt{2/n}$).

\subsection{Methodology for ImageNet Datasets} \label{implementation_details_imagenet}

In our experiments, we used PyTorch 1.10.1 \cite{2019_Paszke_CONF} with four Nvidia GeForce 1080Ti GPUs. All models were trained with the SGD optimizer, set with a momentum of $0.9$ and an initial learning rate of $0.1$, which was reduced by a factor of $0.1$ at epochs $75$, $150$, $225$, $300$, and $375$. A weight decay factor of $1e-4$ was applied, and the batch size was standardized at $128$.

For data augmentation, the training dataset was randomly cropped to $224\times224$, randomly flipped horizontally, and normalized. The validation dataset was resized to $256\times256$, center cropped to $224\times224$, and normalized. Our setup matches the configuration specified in the PyTorch ImageNet training example\footnote{\url{https://github.com/pytorch/examples/blob/main/imagenet}}.

\subsection{Methodology for Cityscapes Datasets} \label{implementation_details_cityscapes}

In the Cityscapes experiments, PyTorch 1.10.1 \cite{2019_Paszke_CONF} and four Nvidia GeForce 1080Ti GPUs were used. Using the MMSegmentation Framework \cite{mmseg2020}, we utilized FCON \cite{2015_Long_CONF} and DeepLab v3 (DLv3) \cite{2017_Chen} on the dataset \cite{2016_Cordts_CONF}, which consists of 2,975 training, 500 validation, and 1,525 testing images across 19 semantic classes. Training involved resizing, random cropping, flipping, photometric distortion, normalization, and padding. Testing employed multi-scale flip augmentation and normalization.

The FCON model, with a ResNet50 backbone, used an Encoder-Decoder architecture with FCONHead as the decode and auxiliary heads. DLv3, also on a ResNet50 backbone, incorporated an Atrous Spatial Pyramid Pooling (ASPP) head with dilations of 1, 12, 24, and 36, and an FCONHead for the auxiliary head. Both models used SyncBN and a dropout ratio of 0.1, with the auxiliary head contributing 40\% to the total loss.

Optimization was via SGD (learning rate 0.01, momentum 0.9, weight decay 0.0005). A polynomial decay learning rate policy was applied (power 0.9, minimum learning rate 1e-4), over 80,000 iterations with checkpoints and evaluations (focusing on mIoU) every 8,000 iterations.

\section{Extended Analysis of FCN Output Weighting on Memory Blocks in ResNet50}
\label{appendix:softmax_activations}

This section builds on the results of the main paper by exploring the weighting mechanism of the FCN output on memory blocks within ResNet50 models using dropblock and dropout2d regularization. We performed a detailed analysis of the softmax activation patterns at the FCN output to further understand this aspect.

\begin{figure*}
	\centering
	% Row 1
	\begin{minipage}[b]{0.30\textwidth}
		\centering
		\includegraphics[width=\textwidth]{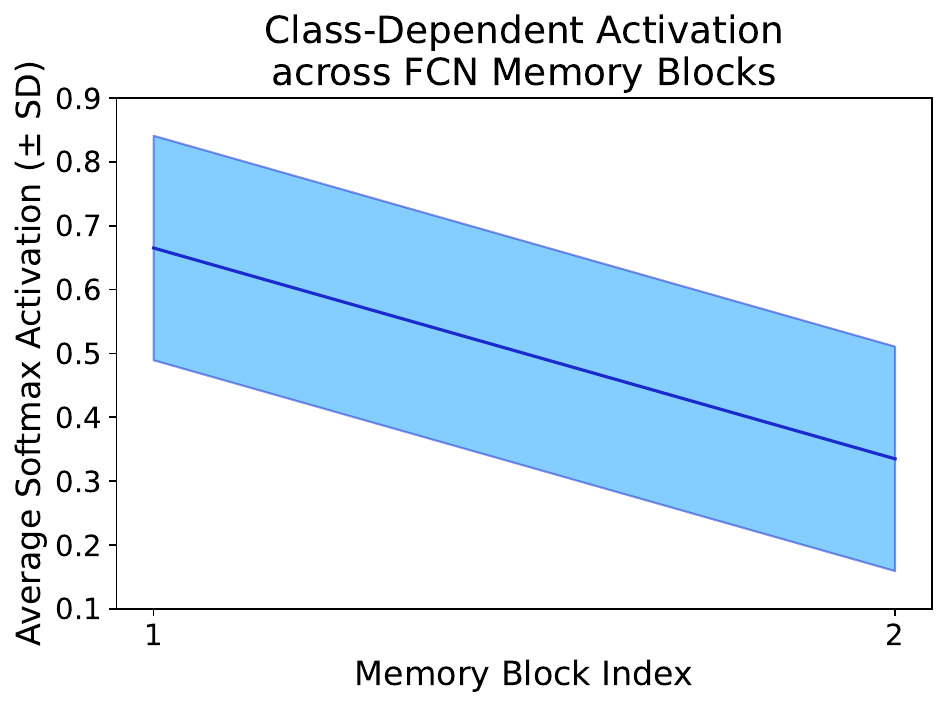}
		Class: All
	\end{minipage}
	\hfill
	\begin{minipage}[b]{0.30\textwidth}
		\centering
		\includegraphics[width=\textwidth]{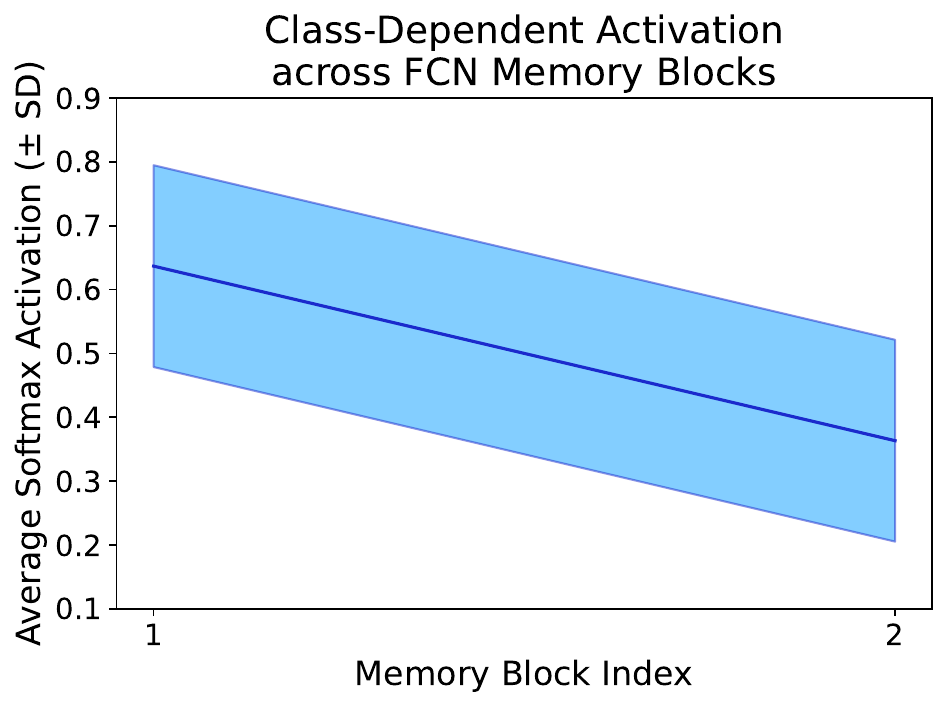}
		Class: Goldfish
	\end{minipage}
	\hfill
	\begin{minipage}[b]{0.30\textwidth}
		\centering
		\includegraphics[width=\textwidth]{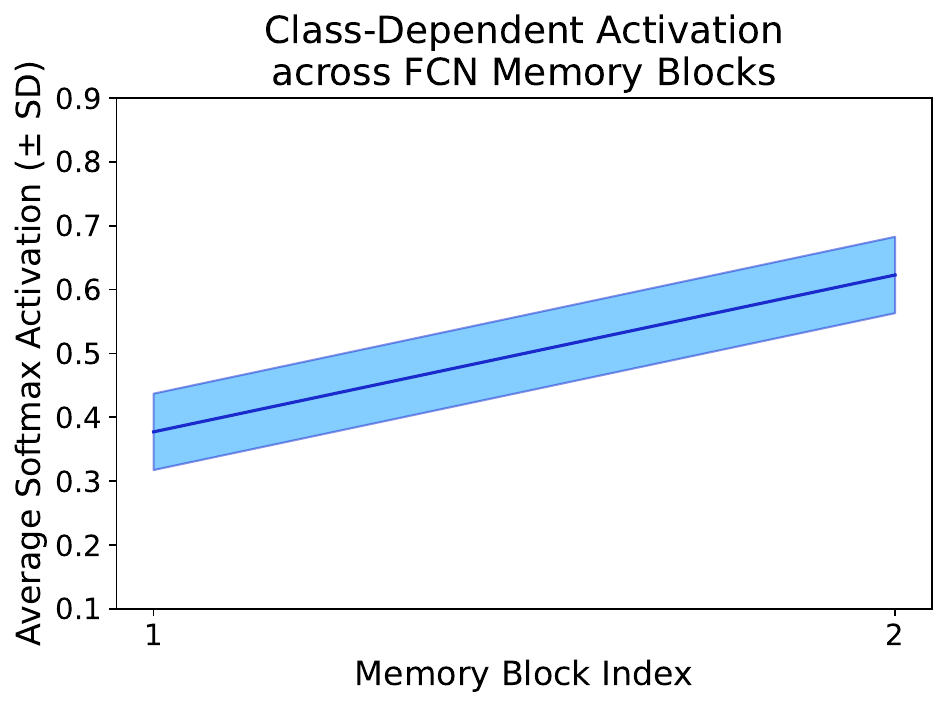}
		Class: Cliff
	\end{minipage}
	
	\medskip % Adds a small space between the rows
	
	% Row 2
	\begin{minipage}[b]{0.30\textwidth}
		\centering
		\includegraphics[width=\textwidth]{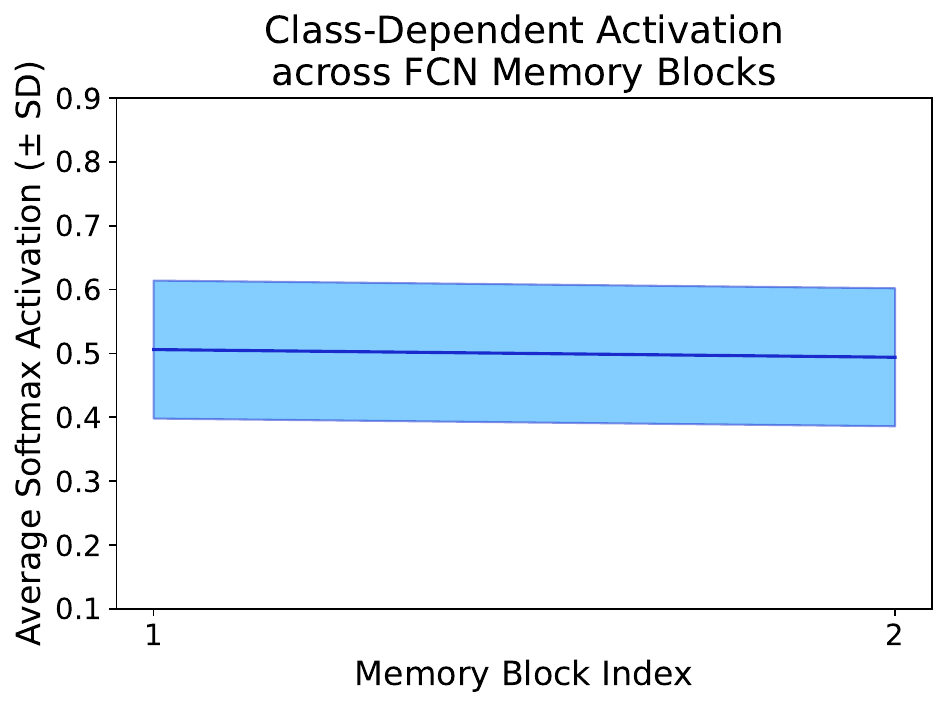}
		Class: Plane
	\end{minipage}
	\hfill
	\begin{minipage}[b]{0.30\textwidth}
		\centering
		\includegraphics[width=\textwidth]{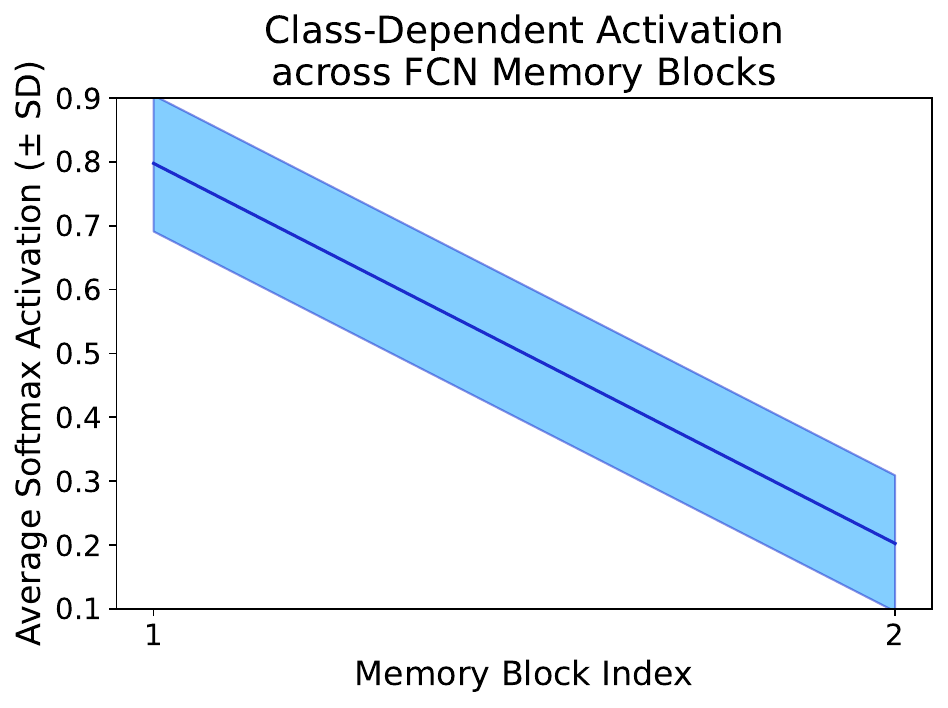}
		Class: Pug
	\end{minipage}
	\hfill
	\begin{minipage}[b]{0.30\textwidth}
		\centering
		\includegraphics[width=\textwidth]{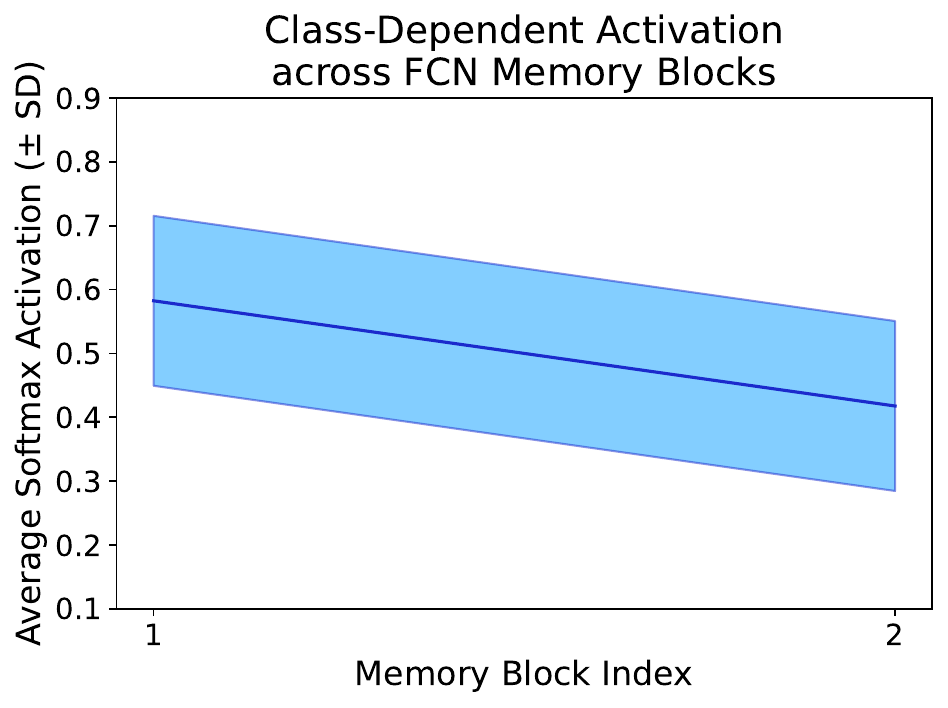}
		Class: Church
	\end{minipage}
	
	\caption{This figure shows the means and standard deviations of softmax activations for ``all'', reflecting the activation across the validation set, and for specific classes, including ``goldfish'', ``cliff'', ``plane'', ``pug'', and ``church''. It illustrates the unique memory usage within a ResNet50 using a SR block with two memory blocks and trained with dropout2d. Notably, ``cliff'' shows a wide range of activation behaviors that differ from the more uniform patterns of ``pug'' and ``church'', highlighting the ability of the FCN to adapt feature processing in a class-dependent manner.}
	\label{fig:activations_1}
\end{figure*}

\begin{figure*}
	\centering
	% Row 1
	\begin{minipage}[b]{0.30\textwidth}
		\centering
		\includegraphics[width=\textwidth]{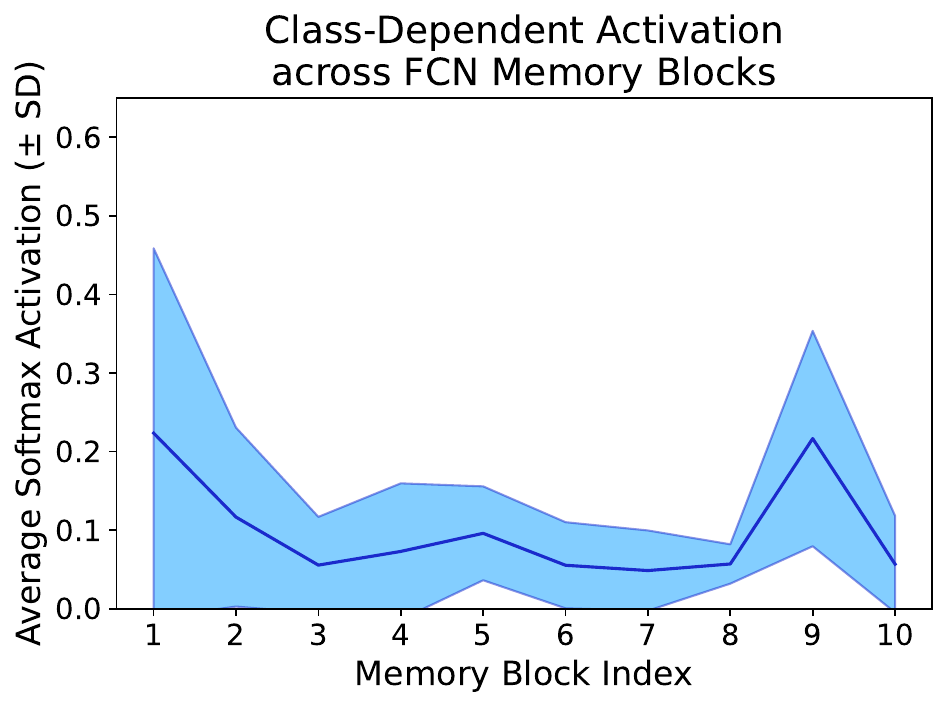}
		Class: All
	\end{minipage}
	\hfill
	\begin{minipage}[b]{0.30\textwidth}
		\centering
		\includegraphics[width=\textwidth]{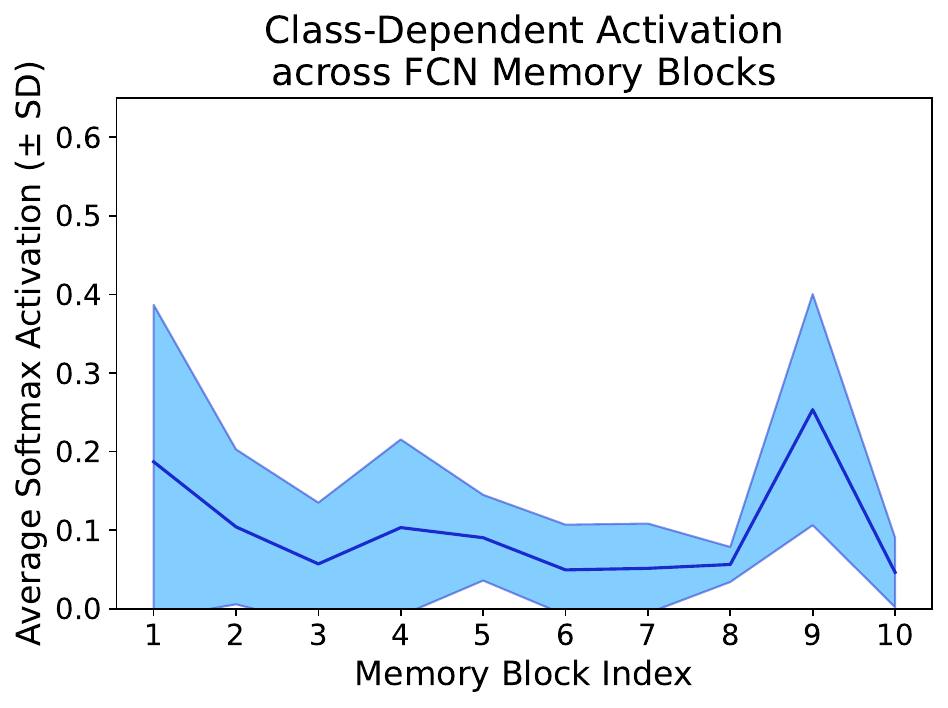}
		Class: Goldfish
	\end{minipage}
	\hfill
	\begin{minipage}[b]{0.30\textwidth}
		\centering
		\includegraphics[width=\textwidth]{Figures/role_of_memory/10_16/ResNet50_dropout2d_o34_bc_ia_3_ch_10_hl_16_s1_cliff_std}
		Class: Cliff
	\end{minipage}
	
	\medskip % Adds a small space between the rows
	
	% Row 2
	\begin{minipage}[b]{0.30\textwidth}
		\centering
		\includegraphics[width=\textwidth]{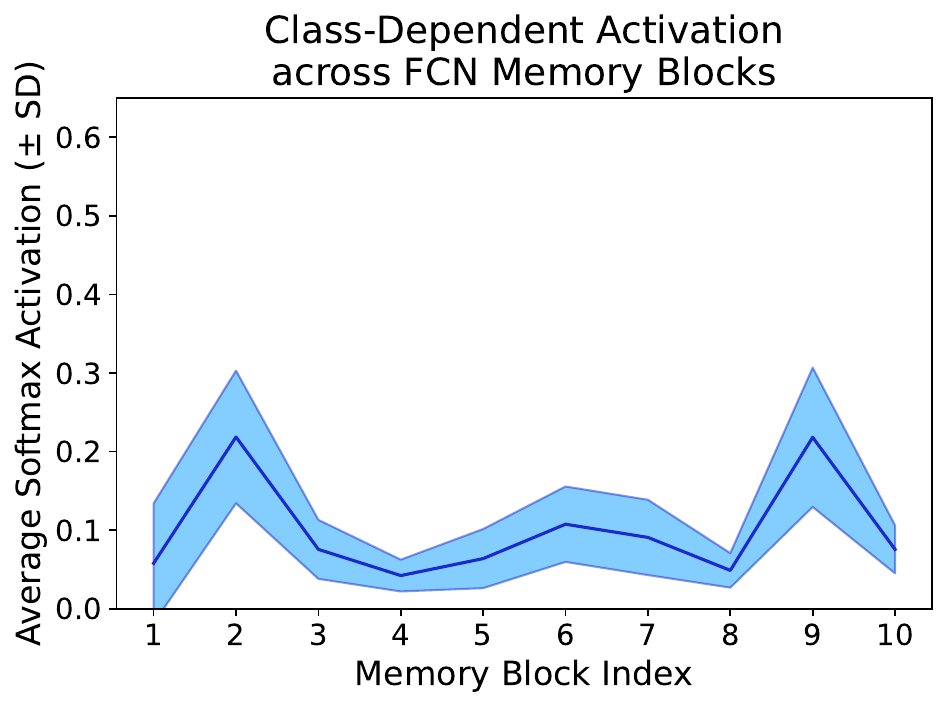}
		Class: Plane
	\end{minipage}
	\hfill
	\begin{minipage}[b]{0.30\textwidth}
		\centering
		\includegraphics[width=\textwidth]{Figures/role_of_memory/10_16/ResNet50_dropout2d_o34_bc_ia_3_ch_10_hl_16_s1_pug_std}
		Class: Pug
	\end{minipage}
	\hfill
	\begin{minipage}[b]{0.30\textwidth}
		\centering
		\includegraphics[width=\textwidth]{Figures/role_of_memory/10_16/ResNet50_dropout2d_o34_bc_ia_3_ch_10_hl_16_s1_church_std}
		Class: Church
	\end{minipage}
	
	\caption{This figure shows the means and standard deviations of softmax activations for ``all'', reflecting the activation across the validation set, and for specific classes, including ``goldfish'', ``cliff'', ``plane'', ``pug'', and ``church''. It illustrates unique memory usage strategies within a ResNet50 using a ten-memory SR block and trained with droput2d. Notably, ``cliff'' exhibits a wide range of activation behaviors that differ from the more uniform patterns of ``pug'' and ``church'', highlighting the ability of the FCN to adapt feature processing in a class-dependent manner.}
	\label{fig:activations_2}
\end{figure*}

\begin{figure*}
	\centering
	% Row 1
	\begin{minipage}[b]{0.3\textwidth}
		\centering
		\includegraphics[width=\textwidth]{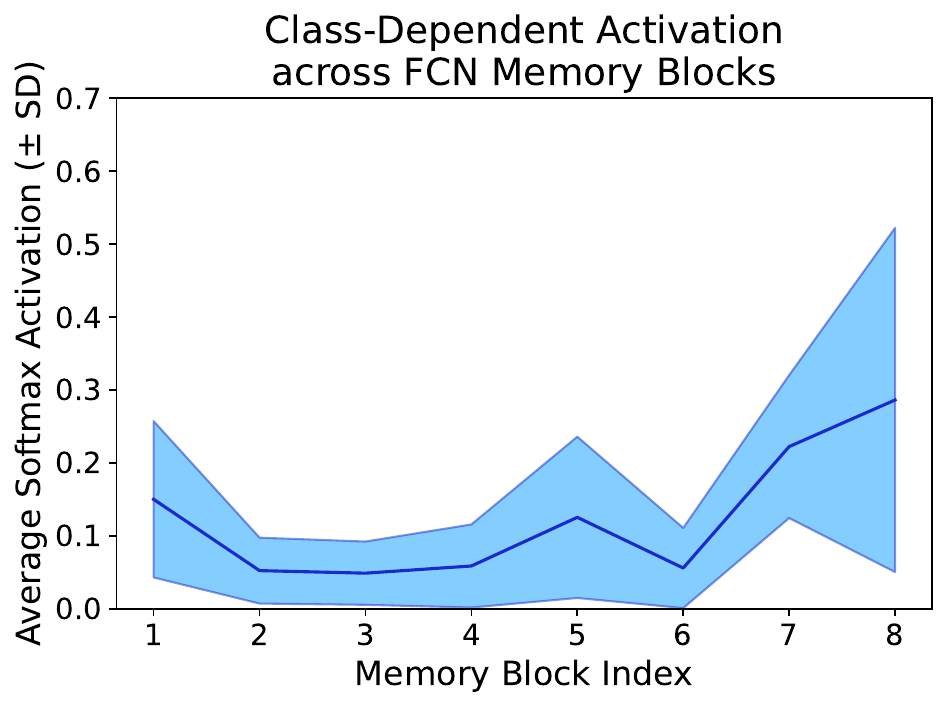}
		Class: All
	\end{minipage}
	\hfill
	\begin{minipage}[b]{0.3\textwidth}
		\centering
		\includegraphics[width=\textwidth]{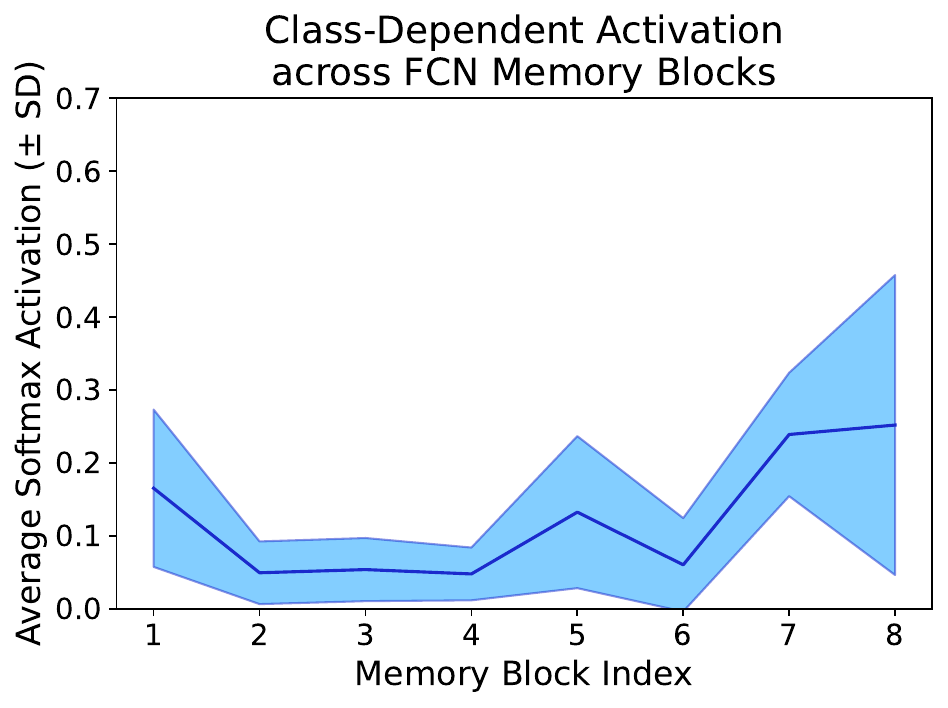}
		Class: Goldfish
	\end{minipage}
	\hfill
	\begin{minipage}[b]{0.3\textwidth}
		\centering
		\includegraphics[width=\textwidth]{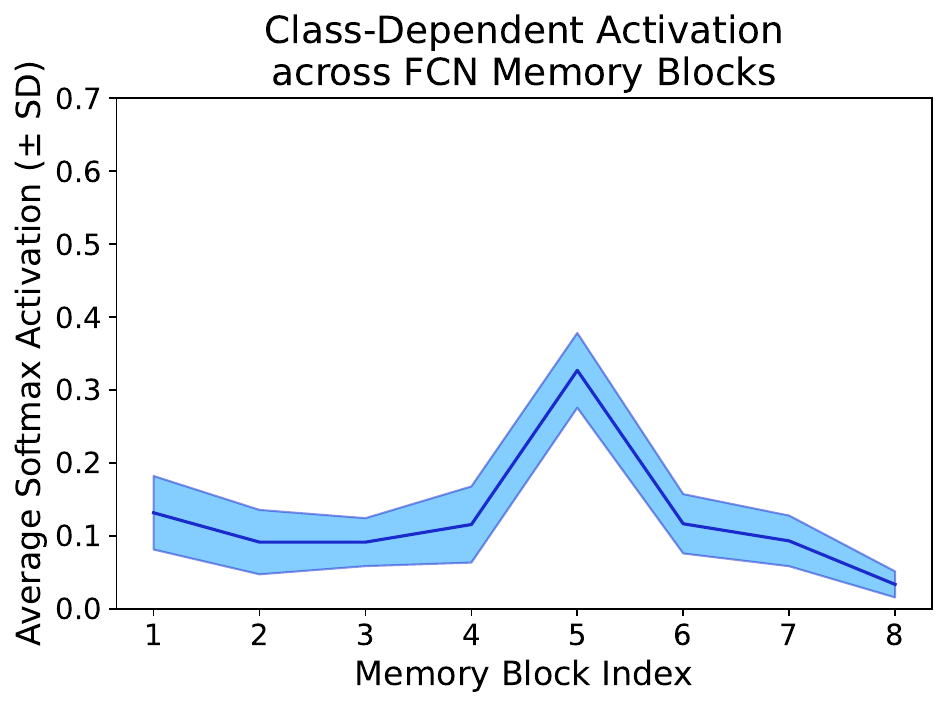}
		Class: Cliff
	\end{minipage}
	
	\medskip % Adds a small space between the rows
	
	% Row 2
	\begin{minipage}[b]{0.3\textwidth}
		\centering
		\includegraphics[width=\textwidth]{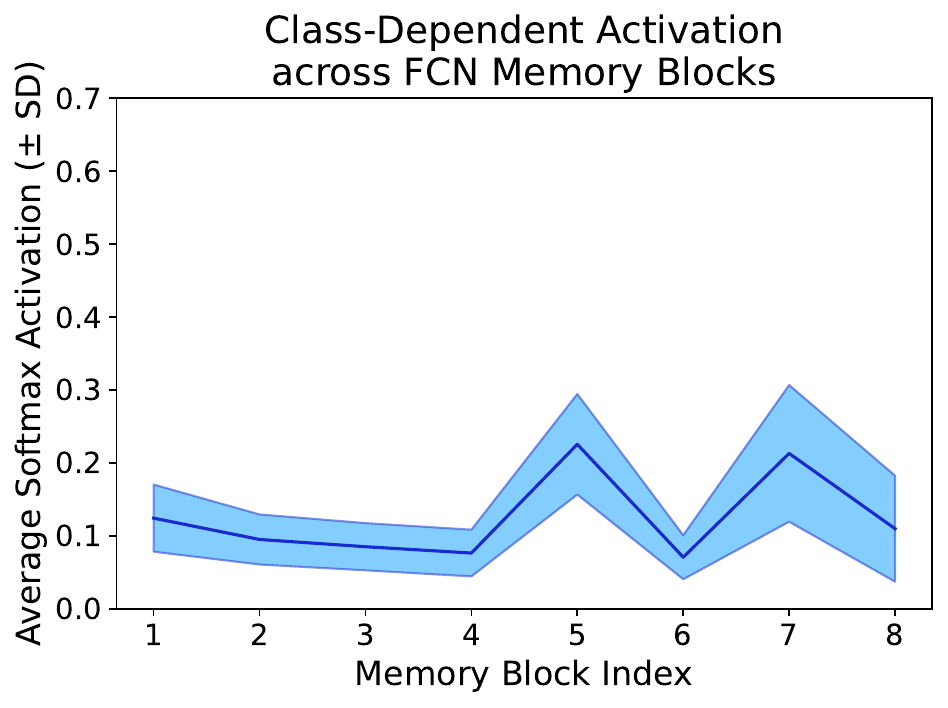}
		Class: Plane
	\end{minipage}
	\hfill
	\begin{minipage}[b]{0.3\textwidth}
		\centering
		\includegraphics[width=\textwidth]{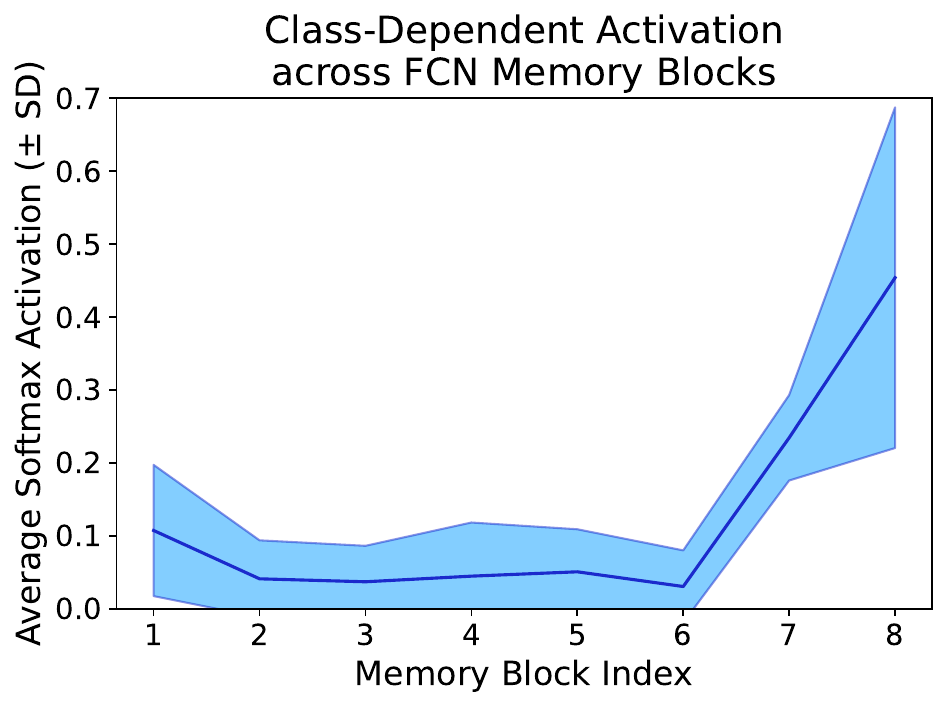}
		Class: Pug
	\end{minipage}
	\hfill
	\begin{minipage}[b]{0.3\textwidth}
		\centering
		\includegraphics[width=\textwidth]{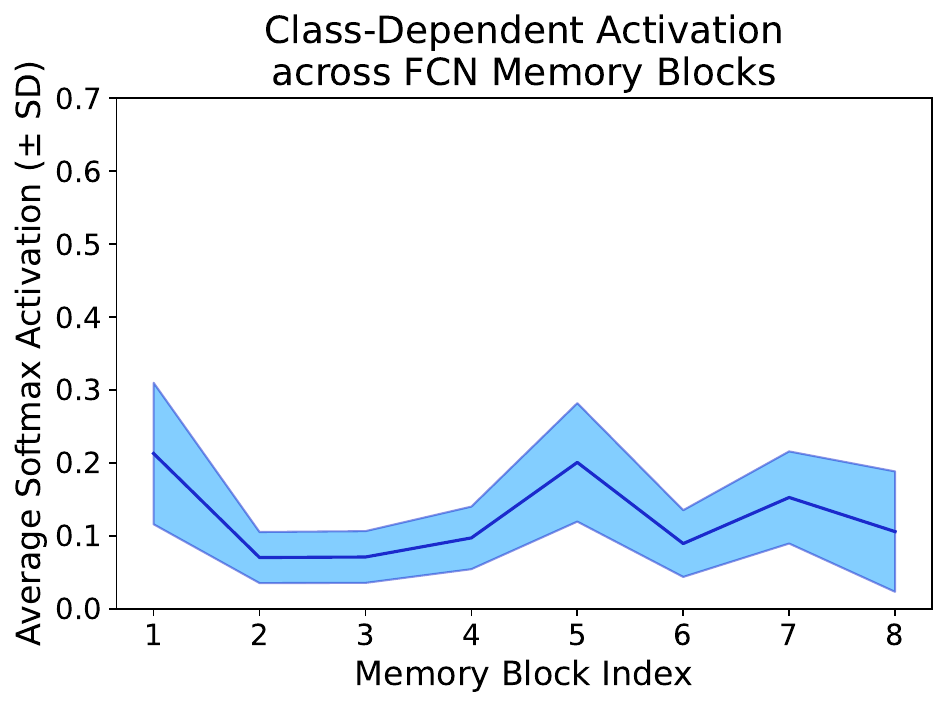}
		Class: Church
	\end{minipage}
	
	\caption{This figure shows the means and standard deviations of softmax activations for ``all'', reflecting the activation across the validation set, and for specific classes, including ``goldfish'', ``cliff'', ``plane'', ``pug'', and ``church''. It illustrates unique memory usage strategies within ResNet50 using an eight-memory SR block and trained with dropblock. Notably, ``cliff'' exhibits a wide range of activation behaviors that differ from the more uniform patterns of ``pug'' and ``church'', highlighting the ability of the FCN to adapt feature processing in a class-dependent manner.}
	\label{fig:activations_3}
\end{figure*}

We selected five classes from the ImageNet dataset to represent a wide range of semantic and visual diversity: ``goldfish'', ``pug'', ``plane'', and ``cliff'' as used also in the SE networks \cite{2018_Hu_CONF}, with the addition of ``church'' to increase the diversity of the dataset. For each class, fifty samples from the validation set were analyzed to compute the mean and standard deviation of softmax activations at the FCN final layer. Figures \ref{fig:activations_1}, \ref{fig:activations_2}, and \ref{fig:activations_3} show these activations for ResNet50 models augmented with SR blocks of different memory sizes and trained with dropout2d and dropblock. They demonstrate the adaptability of the SR block in processing different class features.

We contrast them with the average activations across all 1000 classes using the images from the validation. An important observation is that the SR block does not equal the number of memory blocks to the number of classes; rather, it uses an ensemble of memory blocks. In particular, each class influences the weighting of the memory blocks in a different way, thus controlling the feature enrichment process in a class-specific way.

\end{document}